\documentclass[runningheads]{llncs}

\usepackage{eccv}

\usepackage{eccvabbrv}

\usepackage{graphicx}
\usepackage{booktabs}

\usepackage[accsupp]{axessibility}  

\usepackage{hyperref}

\usepackage{orcidlink}
\usepackage{algorithm}
\usepackage{listings}

\usepackage{multirow}
\usepackage{wrapfig}
\usepackage[most]{tcolorbox}
\usepackage{tcolorbox}

\begin{document}

\title{Beyond Sequential Distance: Inter-Modal Distance Invariant Position Encoding}

\titlerunning{Inter-Modal Distance Invariant Position Encoding}

\vspace{-10pt}
\author{
    Lin Chen\textsuperscript{1,2,3\thanks{Work done during internship at Tencent Hunyuan Team.\\\llap{$\dagger$ }Corresponding author.}}\orcidlink{0000-0001-6965-1459},
    Bolin Ni\textsuperscript{3$\dagger$}\orcidlink{0009-0000-7160-5523},
    Qi Yang\textsuperscript{1,2,3}\orcidlink{0000-0001-8373-6096},
    Zili Wang\textsuperscript{1,2}\orcidlink{0009-0003-0455-7631}, \\
    Kun Ding\textsuperscript{1$\dagger$}\orcidlink{0000-0002-2256-8815},
    Ying Wang\textsuperscript{1}\orcidlink{0000-0003-1385-3224},
    Houwen Peng\textsuperscript{3}\orcidlink{0000-0001-8544-8952} ,
    Shiming Xiang\textsuperscript{1,2}\orcidlink{0000-0002-2089-9733}
}


\institute{
$^1$MAIS, Institute of Automation, Chinese Academy of Sciences, China\\
$^2$University of Chinese Academy of Sciences, China \\
$^3$Tencent Hunyuan Team, China\\
}

\maketitle

\begin{abstract}
Despite the remarkable capabilities of Multimodal Large Language Models (MLLMs), they still suffer from visual fading in long-context scenarios. Specifically, the attention to visual tokens diminishes as the text sequence lengthens, leading to text generation detached from visual constraints. We attribute this degradation to the inherent inductive bias of Multimodal RoPE, which penalizes inter-modal attention as the distance between visual and text tokens increases. To address this, we propose inter-modal \textbf{D}istance \textbf{I}nvariant \textbf{P}osition \textbf{E}ncoding \textbf{(DIPE)}, a simple but effective mechanism that disentangles position encoding based on modality interactions. DIPE retains the natural relative positioning for intra-modal interactions to preserve local structure, while enforcing an anchored perceptual proximity for inter-modal interactions. This strategy effectively mitigates the inter-modal distance-based penalty, ensuring that visual signals remain perceptually consistent regardless of the context length. Experimental results demonstrate that by integrating DIPE with Multimodal RoPE, the model maintains stable visual grounding in long-context scenarios, significantly alleviating visual fading while preserving performance on standard short-context benchmarks. Code is available at \href{https://github.com/lchen1019/DIPE}{\color{magenta}https://github.com/lchen1019/DIPE}.

\keywords{Multimodal Large Language Models \and Position Encoding \and Visual Fading}
\end{abstract}

\section{Introduction}
Multimodal Large Language Models (MLLMs)~\cite{llava,Qwen3-VL,MiniCPM-V,wang2025internvl3_5} have significantly advanced the frontier of machine perception by aligning visual features with the linguistic space of large language models~\cite{touvron2023llama, bai2023qwen, liu2024deepseek, achiam2023gpt}. As model capabilities mature, the research focus has shifted from single-turn visual question answering to complex long-context multimodal understanding~\cite{ge2025v2pe}. In these scenarios, visual signals are no longer inputs requiring only a glance, but are transformed into persistent contexts that must be continuously indexed throughout the generation of hundreds or thousands of tokens.

\begin{figure}
    \centering
    \includegraphics[width=\linewidth]{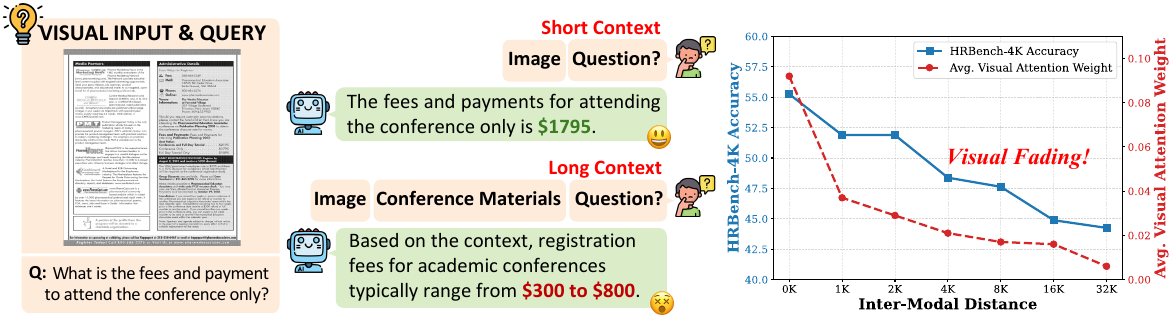}
    \caption{Illustration of the visual fading phenomenon. \textbf{Left}: A qualitative example from DocVQA~\cite{docvqa}. While the model accurately grounds visual evidence in a short-context scenario, it fails to preserve correct attention to the image and generates a wrong answer in a long-context one. \textbf{Right}: Quantitative analysis of visual fading. As the inter-modal distance increases, the proportion of attention allocated to visual tokens exhibits a sharp decay, indicating that the model gradually looks away from the image.}
    \label{fig:intro}
    \vspace{-10pt}
\end{figure}

However, as context length scales, existing MLLMs face a critical limitation, which we term visual fading. As illustrated in Fig.~\ref{fig:intro} (Left), while MLLMs maintain robust attention to visual inputs and successfully ground visual evidence in a short-context scenario, they fail to attend to correct visual cues and generate erroneous predictions in a long-context one. To systematically quantify this phenomenon, we designed a controlled probing setup that incrementally increases the inter-modal distance between visual inputs and text queries. As shown in Fig.~\ref{fig:intro} (Right), our empirical results reveal a significant decay in attention allocated to the image as this distance grows. Consequently, the visual constraints progressively erode during text generation, inevitably increasing the likelihood of generating incorrect responses.

While the natural dilution of softmax attention across long contexts partially accounts for this issue, we argue that the inductive bias from position encoding introduces a non-negligible impact. Current MLLMs typically adopt MRoPE~\cite{Qwen2-VL}, which models the spatial and temporal structures of visual and text tokens in a unified sequential framework by decomposing time, height and width components. However, in such a sequential framework, the relative distance between the initial visual tokens and the newly generated text tokens grows monotonically during autoregressive generation. Crucially, MRoPE retains the long-term decay property~\cite{su2024roformer} from the original RoPE, a mechanism designed to impose a distance penalty for capturing linguistic locality. Consequently, this decay imposes a strong inductive bias that visual information becomes increasingly distant as more text is generated, forcing the model to detach from visual constraints. Furthermore, this distance-based penalty stands in contrast to the sustained visual attention of human cognitive mechanisms~\cite{huang2023review}. Specifically, when a human continuously analyzes an image, the visual reference does not recede into the past like previously spoken words. Instead, it remains right in front of us, regardless of the context length.

To this end, we propose inter-modal \textbf{D}istance \textbf{I}nvariant \textbf{P}osition \textbf{E}ncoding \textbf{(DIPE)}, a simple but effective method that fundamentally redefines how spatial and temporal geometries are modeled across different modalities. Our core insight is that while intra-modal structural integrity must be preserved, the perceptual distance in inter-modal interactions should remain invariant. Specifically, DIPE orthogonally decomposes the attention mechanism: (1) \textbf{Intra-modal Attention}: retains the standard MRoPE to preserve linguistic locality and image's 2D spatial structure. (2) \textbf{Inter-modal Attention}: introduces an anchored query that constrains the perceptual distance between text and visual tokens to a constant. By ensuring visual information remains perceptually proximal to the generation process, DIPE effectively mitigates distance-induced visual fading. DIPE can be seamlessly integrated with RoPE variants in MLLMs without introducing additional parameters. Furthermore, it is compatible with modern efficient attention kernels~\cite{dong2024flexattention} and KV Cache infrastructure.

Experimental results across 19 benchmarks demonstrate that DIPE effectively mitigates the growing visual fading, yielding a 4.10\% average accuracy gain over the MRoPE baseline in long-context scenarios. Crucially, these gains are attained without compromising the performance on standard short-context scenarios. Furthermore, in-depth analyses reveal that DIPE restores visual attention in shallow layers, and consistently maintains significantly higher visual attention weights compared to MRoPE as the context length scales. These findings indicate that DIPE provides a principled and effective approach to mitigating visual fading in MLLMs.

\section{Related Work}
\subsection{Multimodal Large Language Models}
Advancements in large language models~\cite{touvron2023llama, bai2023qwen, liu2024deepseek, achiam2023gpt} have revolutionized multimodal learning by unifying visual and linguistic representations. To achieve modality alignment, early methods employed cross-attention~\cite{alayrac2022flamingo} or Q-Formers~\cite{li2023blip}, while LLaVA~\cite{llava} later introduced a linear projection approach that has become the mainstream paradigm. Recent works have expanded MLLM capabilities to support dense prediction tasks like segmentation~\cite{liu2025seg, liu2026STAMP} and explore lightweight designs~\cite{cai2025llava, chen2026beyond}. Numerous state-of-the-art models have been developed, including the GPT series~\cite{hurst2024gpt} and Gemini~\cite{gemini,gemini_1_5} series, alongside open-source counterparts such as MiniCPM-V series~\cite{MiniCPM-V,yu2025minicpm}, Qwen-VL series~\cite{Qwen-VL,Qwen2-VL,Qwen2.5-VL,Qwen3-VL}, InternVL series~\cite{gao2024mini,chen2024internvl,zhu2025internvl3,wang2025internvl3_5} and LLaVA series~\cite{llavanext,improvedllava,llava}.

\subsection{Position Encoding in MLLMs}
Since the self-attention mechanism is inherently permutation-invariant, incorporating positional information is indispensable for Transformers to model sequence order~\cite{vaswani2017attention}. Existing approaches generally fall into two categories: absolute position encoding (APE)~\cite{vaswani2017attention,dosovitskiy2020image,lan2019albert}, which assigns fixed or learnable vectors to specific token indices, and relative position encoding (RPE)~\cite{shaw2018self,raffel2020exploring,he2020deberta}, which models pairwise distances to capture translation invariance. Rotary Position Embedding (RoPE)~\cite{su2024roformer} has recently emerged as the mainstream foundation for modern LLMs~\cite{touvron2023llama, bai2023qwen}, as it mathematically unifies the benefits of both absolute and relative schemes through rotation matrix while demonstrating superior length extrapolation capabilities~\cite{press2021train}.

In the context of MLLMs, however, applying standard 1D RoPE to flattened visual tokens disrupts spatial-temporal structures. Consequently, recent studies have extended RoPE to higher dimensions, such as 2D-RoPE~\cite{heo2024rotary} for spatial locality in images and 3D MRoPE~\cite{Qwen2-VL} for MLLMs. Building on these foundations, VideoRoPE~\cite{wei2025videorope} and HoPE~\cite{li2025hope} optimize encoding specifically for video dynamics, V2PE~\cite{ge2025v2pe} explores variable visual position encoding, Circle-RoPE~\cite{wang2025circle} projects image token indices onto a ring, MRoPE-I~\cite{huang2025revisiting} revisits frequency allocation strategies and is adopted in Qwen3-VL~\cite{Qwen3-VL}. However, they still penalize inter-modal interactions over distance and trigger visual fading. In contrast, our DIPE ensures constant inter-modal proximity to reflect sustained visual attention. Importantly, rather than competing with these MRoPE variants, DIPE is fully compatible with them, offering a complementary formulation that resolves visual fading while preserving their sophisticated structures.

\section{Preliminaries}
\subsection{Rotary Position Embedding (RoPE)}
RoPE~\cite{su2024roformer} encodes positional information by rotating the query and key vectors within a high-dimensional space. Given a query vector $\boldsymbol{q} \in \mathbb{R}^{d}$ and a key vector $\boldsymbol{k} \in \mathbb{R}^{d}$ corresponding to positions $m$ and $n$, RoPE applies rotation matrices $\boldsymbol{\mathcal{R}}_m \in \mathbb{R}^{d \times d}$ and $\boldsymbol{\mathcal{R}}_n \in \mathbb{R}^{d \times d}$ derived from their absolute position indices. The attention score with RoPE is computed as:
\begin{equation}
    \text{Attn}(\boldsymbol{q}, \boldsymbol{k}) = (\boldsymbol{\mathcal{R}}_m \boldsymbol{q})^\top (\boldsymbol{\mathcal{R}}_n \boldsymbol{k}) = \boldsymbol{q}^\top \boldsymbol{\mathcal{R}}_{n-m} \boldsymbol{k},
\end{equation}
where $\boldsymbol{\mathcal{R}}_{n-m}$ is a matrix that depends solely on the relative distance $n-m$, thereby naturally incorporating relative positional information into the attention mechanism. The rotation matrix $\boldsymbol{\mathcal{R}}_{p}$ (where $p \in \{m, n\}$) is constructed as a block-diagonal matrix, where each block rotates a two-dimensional subspace of the embedding. The rotation angle for the $j$-th subspace is determined by the product of the position index $p$ and a frequency parameter $\theta_j = b^{-2j/d}$, where $b$ is a predefined base constant (e.g., $10000$). This enables the long-term decay property as the relative distance increases.

\subsection{Long-term Decay}
RoPE encodes the relative distance between a query at position $m$ and a key at position $n$ via a rotation matrix $\boldsymbol{\mathcal{R}}_{n-m}$. Su et al.~\cite{su2024roformer} demonstrate that the attention magnitude can be strictly bounded by: 
\begin{equation}
    |\boldsymbol{q}^\top \boldsymbol{\mathcal{R}}_{n-m} \boldsymbol{k}| \leq \mathcal{C} \sum_{j=1}^{d/2} \left| \sum_{l=1}^{j} e^{i\theta_l (n-m)} \right|,
    \label{eq:decay}
\end{equation}
where $\mathcal{C}$ is a bounding constant dependent on the embeddings. As the distance $n-m$ expands, the continuous phase shifts lead to oscillatory cancellation, strictly driving the attention magnitude to decay. This decay injects a strong inductive bias favoring local context, which is essential for linguistic coherence in LLMs. However, it creates a structural conflict in MLLMs. As the response length grows, the relative distance between the newly generated text query and the fixed visual keys increases linearly. Consequently, the attention mechanism mathematically suppresses visual signals as the generation proceeds, causing the model to gradually look away from the image.

\subsection{Multimodal RoPE}
\label{sec:mrope}
\begin{figure}[t]
    \centering
    \includegraphics[width=\linewidth]{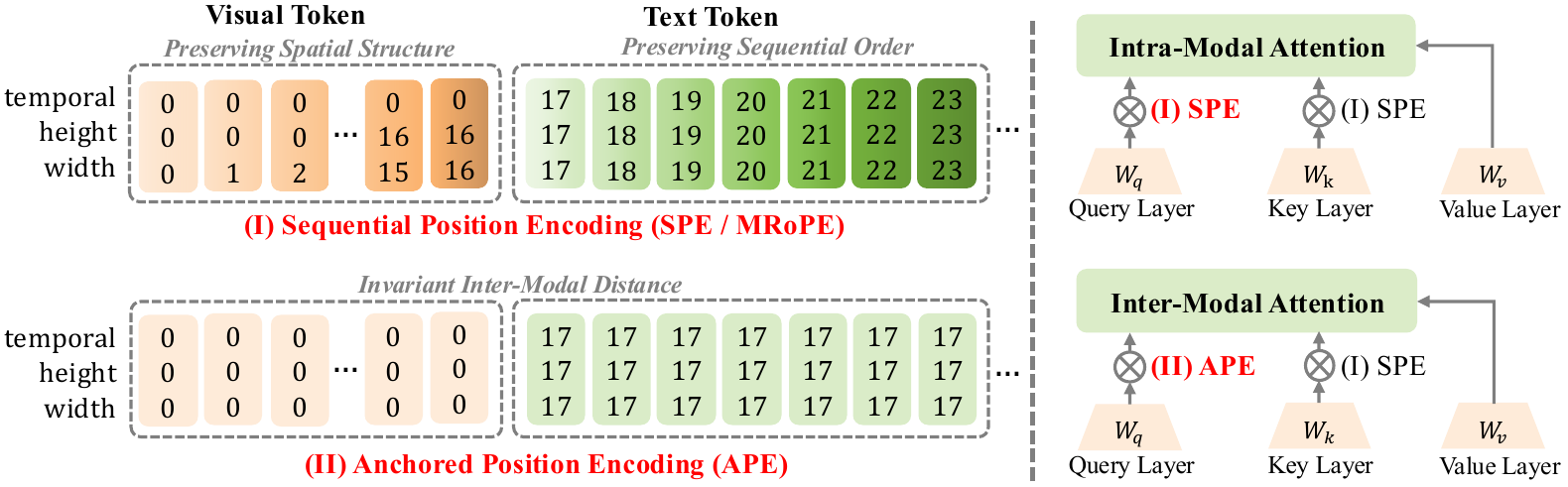}
    \vspace{-10pt}
    \caption{Overview of inter-modal Distance Invariant Position Encoding (DIPE). DIPE mitigates visual fading by disentangling position encoding based on modality interactions. Specifically, intra-modal attention applies sequential position encoding to both queries and keys to preserve spatial and sequential structures. Conversely, inter-modal attention utilizes anchored position encoding for queries alongside sequential position encoding for keys, effectively anchoring the inter-modal perceptual distance.}
    \vspace{-5pt}
    \label{fig:framework}
\end{figure}

While Vanilla RoPE effectively models 1D text sequences, it is suboptimal for MLLMs as it neglects the 2D spatial geometry of images. To address this, Multimodal RoPE (MRoPE)~\cite{Qwen2-VL} decomposes positional information into three components: temporal ($t$), height ($h$) and width ($w$). For the $i$-th token, MRoPE assigns a position tuple $\boldsymbol{p}_i = (t_i, h_i, w_i)$. Given a query or key vector $\mathbf{x} \in \mathbb{R}^d$, MRoPE splits it into three chunks, $\mathbf{x}^{(t)}, \mathbf{x}^{(h)}, \mathbf{x}^{(w)}$, and applies RoPE independently to each chunk using the corresponding position component:
\begin{equation}
\text{MRoPE}(\mathbf{x}, \boldsymbol{p}_i) = \operatorname{concat}
\left(
\boldsymbol{\mathcal{R}}_{t_i}\mathbf{x}^{(t)}, 
\boldsymbol{\mathcal{R}}_{h_i}\mathbf{x}^{(h)}, 
\boldsymbol{\mathcal{R}}_{w_i}\mathbf{x}^{(w)}
\right).
\end{equation}
The long-term decay property described in Eq.~\ref{eq:decay} governs each dimension independently. MRoPE employs distinct positional encoding strategies to preserve the structural integrity of different modalities. 
(1) \textbf{For visual tokens:} to maintain spatial structure, the height and width indices $(h_i, w_i)$ correspond to the grid coordinates of the image patch, while the temporal index $t_i$ remains constant for a given frame. 
(2) \textbf{For text tokens:} to model sequential dependency, text tokens utilize a unified index where the spatial and temporal components are identical (i.e., $t_i = h_i = w_i$). Fig.~\ref{fig:framework} (I) presents an example of MRoPE.

\section{Methodology}

We introduce inter-modal Distance Invariant Position Encoding (DIPE), a simple but effective mechanism designed to eliminate visual fading by reconfiguring inter-modal geometric relationships. Section~\ref{sec:dipe_mechanism} formulates the implementation of DIPE. Section~\ref{sec:implementation} discusses its compatibility with training and inference infrastructures and summarizes the algorithmic implementation.

\subsection{Inter-Modal Distance Invariant Position Encoding}
\label{sec:dipe_mechanism}
Fig.~\ref{fig:framework} illustrates the implementation of DIPE. The core insight of our approach is to decouple position encoding based on the modality relationship between the query and the key. Rather than applying a sequential position index assignment, we introduce sequential position encoding and anchored position encoding to treat intra-modal and inter-modal interactions orthogonally.

\vspace{3pt}
\noindent \textbf{Intra-modal Attention: Preserving Local Structures.}
Intra-modal attention encompasses interactions where both the query and key originate from the same modality (e.g., text-to-text). In these scenarios, maintaining the fidelity of the relative positions is paramount. Text-to-text attention relies on the distance decay of RoPE to model linguistic locality, whereas visual-to-visual attention demands the preservation of 2D spatial geometry for visual coherence.

To preserve these necessary relative distance properties, we utilize Sequential Position Encoding (SPE), which applies standard MRoPE position tuples, denoted as $\boldsymbol{p}^{\text{spe}}$, to both queries and keys. The intra-modal attention score between a query vector $\boldsymbol{q}$ at position $m$ and a key vector $\boldsymbol{k}$ at position $n$ is computed as:
\begin{equation}
    \operatorname{Attn}_{\text{intra}}(\boldsymbol{q}, \boldsymbol{k}) = (\boldsymbol{\mathcal{R}}_{\boldsymbol{p}^{\text{spe}}_m} \boldsymbol{q})^\top (\boldsymbol{\mathcal{R}}_{\boldsymbol{p}^{\text{spe}}_n} \boldsymbol{k}),
\end{equation}
where $\text{Modality}(\boldsymbol{q}) = \text{Modality}(\boldsymbol{k})$. This formulation strictly retains the structural integrity of both linguistic and visual domains.

\vspace{3pt}
\noindent \textbf{Inter-modal Attention: Invariant Inter-Modal Distance.}
Inter-modal attention specifically refers to queries and keys from different modalities. This is where the distance decay described in Eq.~\ref{eq:decay} becomes detrimental. 

To counteract this, we propose the Anchored Position Encoding (APE). For any inter-modal interaction, we assign anchored position indices to the queries, denoted as $\boldsymbol{p}^{\text{ape}}$. Specifically, for each partitioned modality segment, we extract the position index of its first token. This initial position serves as a unified anchor for all tokens within that entire segment when they act as queries. By applying APE to the query while maintaining SPE for the key, we ensure that the inter-modal relative distance remains invariant with respect to the autoregressive generation steps. The inter-modal attention score between a query vector $\boldsymbol{q}$ at position $m$ and a key vector $\boldsymbol{k}$ at position $n$ is computed as:
\begin{equation}
    \text{Attn}_{\text{inter}}(\boldsymbol{q}, \boldsymbol{k}) = (\boldsymbol{\mathcal{R}}_{\boldsymbol{p}^{\text{ape}}_m} \boldsymbol{q})^\top (\boldsymbol{\mathcal{R}}_{\boldsymbol{p}^{\text{spe}}_n} \boldsymbol{k}),
\end{equation}
where $\text{Modality}(\boldsymbol{q}) \neq \text{Modality}(\boldsymbol{k})$. This bidirectional anchoring ensures that the perceptual distance between visual and text tokens remains constant.

\vspace{3pt}
\noindent \textbf{Implementation of Position Design in DIPE.}
Algorithm~\ref{alg:index_code} presents the implementation details of the sequential and anchored position index allocation logic in DIPE. For clarity, the input of the function is defined as a list of sequentially partitioned modality segments. SPE naturally adopts the position allocation strategy of MRoPE to maintain local structures. Meanwhile, APE extracts the starting $(t, h, w)$ coordinates of each segment and broadcasts them across all tokens within that segment.

\noindent \textbf{Robustness to Multi-Image and Interleaved Contexts.}
The formulation of DIPE is inherently designed for interleaved and multi-image contexts. By treating each modality segment as a positional unit, DIPE assigns a unique anchor to each segment based on its relative temporal onset. This unified approach ensures that the macroscopic sequence order is strictly preserved, while the inter-modal perceptual distance for each specific visual entity remains invariant. Experimental results to support this are provided in Sec.~\ref{sec:interleave}.

\begin{algorithm}[t]
\caption{PyTorch-style Pseudocode for Position Design in DIPE.}
\label{alg:index_code}
\newcommand{\algcomment}[1]{\hfill\textcolor{codeblue}{\# #1}}
\definecolor{codeblue}{rgb}{0.25,0.5,0.5}
\lstset{
  backgroundcolor=\color{white},
  basicstyle=\fontsize{7.5pt}{8.5pt}\ttfamily\selectfont,
  columns=fullflexible,
  breaklines=true,
  frame=none, 
  showstringspaces=false,
  keepspaces=true,
  commentstyle=\color{codeblue},
  keywordstyle=\bfseries,
  belowskip=-1pt,
}
\vspace{-5pt}
\begin{lstlisting}[language=python]
def generate_dipe_indices(
    modality_segments: List[Tensor], # partitioned segments (e.g., [Txt, Img1, Txt, Img2...])
):
    pos_spe, pos_ape = [], []
    seq_offset = 0
    for seg in modality_segments:
        # 1. SPE: adopt indices from MRoPE, and can be replaced with MRoPE variants
        spe_seg = get_mrope_ids(seg, seq_offset)
        
        # 2. APE: Broadcast the first token's (T,H,W) across the entire segment
        ape_seg = torch.full_like(spe_seg, spe_seg[:, [0]])
                
        # 3. Update seq_offset and save results for the next segment
        seq_offset = spe_seg.max() + 1 # strategy from MRoPE
        pos_spe.append(spe_seg), pos_ape.append(ape_seg)
        
    return torch.cat(pos_spe, dim=-1), torch.cat(pos_ape, dim=-1)
\end{lstlisting}
\end{algorithm}

\subsection{Training and Inference Infrastructure for DIPE}
\label{sec:implementation}
While DIPE theoretically necessitates decoupling the attention computation into intra-modal and inter-modal blocks, we demonstrate that this can be implemented efficiently within modern LLM infrastructures. Our design is compatible with modern efficient attention kernels such as FlexAttention~\cite{dong2024flexattention} and standard KV cache.

\noindent \textbf{Decoupled Attention with LogSumExp Fusion.}
Since DIPE applies different rotations to the query depending on the target key, we employ a split-kernel strategy. Specifically, we execute two modality-masked attention kernels. (1) Inter-modal kernel: computes inter-modal attention using the anchored query and sequential key. (2) Intra-modal kernel: computes intra-modal attention using the sequential query and sequential key.

In our implementation, this split attention is supported by efficient mask-aware kernels such as FlexAttention~\cite{dong2024flexattention}. To merge the outputs, we utilize the LogSumExp ($\log \sum \exp(\cdot)$) statistic returned by attention kernels. Given outputs $\mathbf{O}_1, \mathbf{O}_2$ and their corresponding LogSumExp normalization factors $\ell_1, \ell_2$, the merged output $\mathbf{O}$ is derived as:
\begin{equation}
    \mathbf{O} = \frac{e^{\ell_1} \mathbf{O}_1 + e^{\ell_2} \mathbf{O}_2}{e^{\ell_1} + e^{\ell_2}} = \sigma(\ell_1 - \ell_2) \mathbf{O}_1 + \sigma(\ell_2- \ell_1) \mathbf{O}_2,
\end{equation}
where $\sigma$ is the sigmoid function. The derivation is provided in Appendix~\ref{appendix:proof-lse}.

\begin{algorithm}[t]
\caption{PyTorch-style Pseudocode for Attention With DIPE.}
\label{alg:va_code}
\newcommand{\algcomment}[1]{\hfill\textcolor{codeblue}{\# #1}}
\definecolor{codeblue}{rgb}{0.25,0.5,0.5}
\lstset{
  backgroundcolor=\color{white},
  basicstyle=\fontsize{7.5pt}{8.5pt}\ttfamily\selectfont,
  columns=fullflexible,
  breaklines=true,
  frame=none, 
  showstringspaces=false,
  keepspaces=true,
  commentstyle=\color{codeblue},
  keywordstyle=\bfseries,
  belowskip=-1pt,
}
\vspace{-5pt}
\begin{lstlisting}[language=python]
def attn_forward_with_dipe(
    hidden_states: Tensor,       # [seq_len, dim]
    pos_spe: Tensor,             # sequential position indices, [3, seq_len]
    pos_ape: Tensor,             # anchored position indices, [3, seq_len]
    intra_attn_mask: Tensor,     # attention mask for intra-modal, [seq_len, seq_len]
    inter_attn_mask: Tensor,     # attention mask for inter-modal, [seq_len, seq_len]
    past_key_values: Optional[Cache] = None,  # KV Cache for inference
):
    # 1. Linear projections of query, key and value
    q, k, v = q_proj(hidden_states), k_proj(hidden_states), v_proj(hidden_states)
    
    # 2. Apply position encoding
    q_spe = apply_mrope(q, pos_spe)
    q_ape = apply_mrope(q, pos_ape)
    k = apply_mrope(k, pos_spe)

    # 3. Update KV cache if inference
    if past_key_values is not None:
        k, v = past_key_values.update(k, v)

    # 4. Decoupled attention with intra-modal and inter-modal masks
    o_intra, lse_intra = flex_attn_func(q_spe, k, v, intra_attn_mask)
    o_inter, lse_inter = flex_attn_func(q_ape, k, v, inter_attn_mask)

    # 5. Merge via LogSumExp (Eq.6)
    alpha = sigmoid(lse_intra - lse_inter)
    output = alpha * o_intra + (1 - alpha) * o_inter
    
    return output
\end{lstlisting}
\end{algorithm}

\vspace{3pt}
\noindent \textbf{Seamless Integration with KV Cache.}
DIPE is strictly a query-side intervention. It requires creating two views of the query vector during the forward pass. Critically, it does not alter the content of previous keys and values. This allows DIPE to be deployed on top of pre-computed KV caches without requiring cache re-indexing or memory layout modifications.

\vspace{3pt}
\noindent \textbf{Overall Implementation.}
The execution logic of attention with DIPE is summarized in Algorithm~\ref{alg:va_code}. Integrating DIPE into a standard forward pass involves three modifications: First, generating dual query views by applying SPE and APE, respectively. Second, executing attention computations for each query view using modality-specific masks. Finally, fusing the resulting output vectors via the LogSumExp trick, ensuring mathematically rigorous normalization across the split attention kernels.

\section{Experimental Results}
\subsection{Experimental Settings}
\noindent \textbf{Model Architecture.}
We employ Qwen2.5-3B~\cite{qwen2.5} as our primary LLM backbone, supplemented by Qwen2.5-0.5B and Qwen3-1.7B~\cite{yang2025qwen3} for scalability analysis. For visual encoding, we primarily utilize SigLIP2-SO400M~\cite{tschannen2025siglip} with the NaFlex variant to support dynamic resolution. The visual features are projected using a simple two-layer MLP with GELU activation.

\noindent \textbf{Training Recipe.}
Following established paradigms~\cite{llava}, we adopt a two-stage training pipeline. The first stage focuses on vision-language alignment by freezing both the vision encoder and the language model to initialize the visual projector. The second stage performs instruction tuning by unfreezing the LLM. We leverage the LLaVA-Pretrain-558K~\cite{llava} and LLaVA-NeXT-779K~\cite{llavanext} datasets for these two stages, respectively. Detailed configurations are listed in Appendix~\ref{appendix:hyperparams}.

\noindent \textbf{Evaluation Protocols.}
To comprehensively assess model performance and its robustness against visual fading, we evaluate our approach on 19 widely adopted benchmarks spanning three core domains: perception, document understanding and general visual question answering (VQA). Specifically, the perception benchmarks include CountBench~\cite{countbench}, HRBench~\cite{hrbench}, V$^*$~\cite{vstar}, POPE~\cite{pope} and BLINK~\cite{blink}. The document understanding benchmarks consist of ChartQA~\cite{chartqa}, DocVQA~\cite{docvqa}, TextVQA~\cite{textvqa}, AI2D~\cite{ai2d}, InfoVQA~\cite{infovqa} and OCRBench~\cite{ocrbench}. For general VQA, we evaluate on RealWorldQA~\cite{realworldqa}, MMStar~\cite{mmstar}, MMBench~\cite{mmbench}, MMVP~\cite{mmvp}, MathVision~\cite{mathvision} and MathVista~\cite{mathvista}. We evaluate these benchmarks under two distinct protocols: Short-Context VQA and Long-Context VQA.

\textbf{Short-Context VQA} protocol strictly follows the standard evaluation settings of the original benchmarks, where the question immediately follows the image. Its primary objective is to verify whether our proposed DIPE preserves the fundamental capabilities of the MLLM without compromising accuracy in standard short-context scenarios.

\textbf{Long-Context VQA} protocol is a controlled diagnostic stress test designed specifically to simulate and quantify the visual fading phenomenon, rather than a conventional long-context benchmark. It systematically expands the relative distance between the visual context and the subsequent task. We source text from the standard distractors corpus~\cite{gao2025u, sangpark2025dnotitianiah, li2024needlebench} and insert them directly between the visual input and the question to simulate long contexts and increase the distance between text and images. We intentionally employ these distractors rather than task-related content to isolate the effect of distance on visual attention, ensuring that the evaluation focuses purely on visual attention degradation without the confounding factors of semantic reasoning over related text. By progressively varying the distractor length from 1K to 32K tokens, this protocol evaluates the model's robustness against visual attention degradation.

Additionally, we employ \textbf{MM-NIAH}~\cite{niah} to evaluate standard long-context capabilities in complex real-world scenarios to complement the Long-Context VQA protocol. This benchmark validates the model's ability to reason over specific needles embedded within extensive interleaved multimodal contexts.

\begin{table}[t!]
    \centering
    \small
    \setlength{\tabcolsep}{3pt}
    \renewcommand{\arraystretch}{1.03}
    \caption{Performance comparison under the Long-Context VQA protocol. We evaluate Vanilla RoPE~\cite{su2024roformer}, MRoPE~\cite{Qwen2-VL} and MRoPE-I~\cite{huang2025revisiting} baselines alongside their DIPE-enhanced counterparts.}
    \vspace{-10pt}
    \label{tab:performance_evaluation}
    \resizebox{\columnwidth}{!}{ 
    \begin{tabular}{ll llllll}
        \toprule
        \multirow{2}{*}{\textbf{Capability}} & \multirow{2}{*}{\textbf{Benchmark}} & \multicolumn{2}{c}{\textbf{Vanilla RoPE}} & \multicolumn{2}{c}{\textbf{MRoPE}} & \multicolumn{2}{c}{\textbf{MRoPE-I}} \\
        \cmidrule(lr){3-4} \cmidrule(lr){5-6} \cmidrule(lr){7-8}
        & & Base & +DIPE & Base & +DIPE & Base & +DIPE \\
        \midrule
        
        \multirow{6}{*}{Perception} 
        & HRBench-4K~\cite{hrbench} & 50.75 & 51.00{\scriptsize \textcolor[rgb]{0.2,0.6,0.2}{(+0.25)}} & 47.63 & 54.50{\scriptsize \textcolor[rgb]{0.2,0.6,0.2}{(+6.87)}} & 51.62 & 52.12{\scriptsize \textcolor[rgb]{0.2,0.6,0.2}{(+0.50)}} \\
        & HRBench-8K~\cite{hrbench} & 42.63 & 44.12{\scriptsize \textcolor[rgb]{0.2,0.6,0.2}{(+1.49)}} & 42.25 & 45.13{\scriptsize \textcolor[rgb]{0.2,0.6,0.2}{(+2.88)}} & 43.75 & 43.88{\scriptsize \textcolor[rgb]{0.2,0.6,0.2}{(+0.13)}} \\
        & V$^*$~\cite{vstar} & 47.12 & 48.17{\scriptsize \textcolor[rgb]{0.2,0.6,0.2}{(+1.05)}} & 49.21 & 50.26{\scriptsize \textcolor[rgb]{0.2,0.6,0.2}{(+1.05)}} & 50.79 & 47.64{\scriptsize \textcolor{red}{(-3.15)}} \\
        & POPE~\cite{pope} & 78.49 & 81.37{\scriptsize \textcolor[rgb]{0.2,0.6,0.2}{(+2.88)}} & 74.50 & 85.58{\scriptsize \textcolor[rgb]{0.2,0.6,0.2}{(+11.08)}} & 75.13 & 83.72{\scriptsize \textcolor[rgb]{0.2,0.6,0.2}{(+8.59)}} \\
        & CountBench~\cite{countbench} & 70.26 & 75.15{\scriptsize \textcolor[rgb]{0.2,0.6,0.2}{(+4.89)}} & 66.60 & 75.97{\scriptsize \textcolor[rgb]{0.2,0.6,0.2}{(+9.37)}} & 67.62 & 74.95{\scriptsize \textcolor[rgb]{0.2,0.6,0.2}{(+7.33)}} \\
        & BLINK$_{\mathrm{val}}$~\cite{blink} & 40.93 & 39.19{\scriptsize \textcolor{red}{(-1.74)}} & 39.87 & 41.93{\scriptsize \textcolor[rgb]{0.2,0.6,0.2}{(+2.06)}} & 40.29 & 42.45{\scriptsize \textcolor[rgb]{0.2,0.6,0.2}{(+2.16)}} \\
        \midrule

        \multirow{6}{*}{\shortstack{Document \\ Understanding}} 
        & ChartQA$_{\mathrm{test}}$~\cite{chartqa} & 44.20 & 46.40{\scriptsize \textcolor[rgb]{0.2,0.6,0.2}{(+2.20)}} & 40.44 & 45.16{\scriptsize \textcolor[rgb]{0.2,0.6,0.2}{(+4.72)}} & 44.92 & 47.60{\scriptsize \textcolor[rgb]{0.2,0.6,0.2}{(+2.68)}} \\
        & DocVQA$_{\mathrm{val}}$~\cite{docvqa} & 38.38 & 40.18{\scriptsize \textcolor[rgb]{0.2,0.6,0.2}{(+1.80)}} & 33.25 & 38.55{\scriptsize \textcolor[rgb]{0.2,0.6,0.2}{(+5.30)}} & 36.01 & 39.90{\scriptsize \textcolor[rgb]{0.2,0.6,0.2}{(+3.89)}} \\
        & AI2D$_{\mathrm{test}}$~\cite{ai2d} & 65.06 & 66.48{\scriptsize \textcolor[rgb]{0.2,0.6,0.2}{(+1.42)}} & 65.32 & 67.39{\scriptsize \textcolor[rgb]{0.2,0.6,0.2}{(+2.07)}} & 66.26 & 67.39{\scriptsize \textcolor[rgb]{0.2,0.6,0.2}{(+1.13)}} \\
        & TextVQA$_{\mathrm{val}}$~\cite{textvqa} & 48.17 & 51.60{\scriptsize \textcolor[rgb]{0.2,0.6,0.2}{(+3.43)}} & 46.61 & 50.03{\scriptsize \textcolor[rgb]{0.2,0.6,0.2}{(+3.42)}} & 49.75 & 52.34{\scriptsize \textcolor[rgb]{0.2,0.6,0.2}{(+2.59)}} \\
        & InfoVQA$_{\mathrm{val}}$~\cite{infovqa} & 21.92 & 21.91{\scriptsize \textcolor{red}{(-0.01)}} & 23.60 & 21.84{\scriptsize \textcolor{red}{(-1.76)}} & 22.34 & 23.26{\scriptsize \textcolor[rgb]{0.2,0.6,0.2}{(+0.92)}} \\
        & OCRBench~\cite{ocrbench} & 23.90 & 25.20{\scriptsize \textcolor[rgb]{0.2,0.6,0.2}{(+1.30)}} & 22.50 & 23.40{\scriptsize \textcolor[rgb]{0.2,0.6,0.2}{(+0.90)}} & 22.50 & 25.30{\scriptsize \textcolor[rgb]{0.2,0.6,0.2}{(+2.80)}} \\
        \midrule

        \multirow{7}{*}{General VQA}
        & RealWorldQA~\cite{realworldqa} & 54.51 & 55.29{\scriptsize \textcolor[rgb]{0.2,0.6,0.2}{(+0.78)}} & 52.81 & 58.82{\scriptsize \textcolor[rgb]{0.2,0.6,0.2}{(+6.01)}} & 55.95 & 56.08{\scriptsize \textcolor[rgb]{0.2,0.6,0.2}{(+0.13)}} \\
        & MMStar~\cite{mmstar} & 38.53 & 40.20{\scriptsize \textcolor[rgb]{0.2,0.6,0.2}{(+1.67)}} & 38.80 & 41.67{\scriptsize \textcolor[rgb]{0.2,0.6,0.2}{(+2.87)}} & 38.47 & 40.73{\scriptsize \textcolor[rgb]{0.2,0.6,0.2}{(+2.26)}} \\
        & MMBenchV1.1-EN$_{\mathrm{dev}}$~\cite{mmbench} & 53.64 & 57.89{\scriptsize \textcolor[rgb]{0.2,0.6,0.2}{(+4.25)}} & 50.00 & 58.75{\scriptsize \textcolor[rgb]{0.2,0.6,0.2}{(+8.75)}} & 52.86 & 56.89{\scriptsize \textcolor[rgb]{0.2,0.6,0.2}{(+4.03)}} \\
        & MMBenchV1.1-CN$_{\mathrm{dev}}$~\cite{mmbench} & 52.01 & 56.50{\scriptsize \textcolor[rgb]{0.2,0.6,0.2}{(+4.49)}} & 49.30 & 56.66{\scriptsize \textcolor[rgb]{0.2,0.6,0.2}{(+7.36)}} & 52.79 & 54.95{\scriptsize \textcolor[rgb]{0.2,0.6,0.2}{(+2.16)}} \\
        & MMVP~\cite{mmvp} & 58.67 & 61.33{\scriptsize \textcolor[rgb]{0.2,0.6,0.2}{(+2.66)}} & 56.67 & 60.00{\scriptsize \textcolor[rgb]{0.2,0.6,0.2}{(+3.33)}} & 60.67 & 61.33{\scriptsize \textcolor[rgb]{0.2,0.6,0.2}{(+0.66)}} \\
        & MathVision$_{\mathrm{mini}}$~\cite{mathvision} & 19.74 & 22.70{\scriptsize \textcolor[rgb]{0.2,0.6,0.2}{(+2.96)}} & 18.09 & 19.08{\scriptsize \textcolor[rgb]{0.2,0.6,0.2}{(+0.99)}} & 19.08 & 18.75{\scriptsize \textcolor{red}{(-0.33)}} \\
        & MathVista$_{\mathrm{mini}}$~\cite{mathvista} & 30.90 & 33.20{\scriptsize \textcolor[rgb]{0.2,0.6,0.2}{(+2.30)}} & 31.70 & 32.20{\scriptsize \textcolor[rgb]{0.2,0.6,0.2}{(+0.50)}} & 32.70 & 32.40{\scriptsize \textcolor{red}{(-0.30)}} \\ \midrule

        \multirow{5}{*}{Overall} 
        & Perception & 55.03 & 56.50{\scriptsize \textcolor[rgb]{0.2,0.6,0.2}{(+1.47)}} & 53.34 & 58.90{\scriptsize \textcolor[rgb]{0.2,0.6,0.2}{(+5.56)}} & 54.87 & 57.46{\scriptsize \textcolor[rgb]{0.2,0.6,0.2}{(+2.59)}} \\
        & Document & 40.27 & 41.96{\scriptsize \textcolor[rgb]{0.2,0.6,0.2}{(+1.69)}} & 38.62 & 41.06{\scriptsize \textcolor[rgb]{0.2,0.6,0.2}{(+2.44)}} & 40.30 & 42.63{\scriptsize \textcolor[rgb]{0.2,0.6,0.2}{(+2.33)}} \\
        & General & 44.00 & 46.73{\scriptsize \textcolor[rgb]{0.2,0.6,0.2}{(+2.73)}} & 42.48 & 46.74{\scriptsize \textcolor[rgb]{0.2,0.6,0.2}{(+4.26)}} & 44.65 & 45.88{\scriptsize \textcolor[rgb]{0.2,0.6,0.2}{(+1.23)}} \\
        \cmidrule{2-8}
        & All & 46.31 & 48.31{\scriptsize \textcolor[rgb]{0.2,0.6,0.2}{(+2.00)}} & 44.69 & 48.79{\scriptsize \textcolor[rgb]{0.2,0.6,0.2}{(+4.10)}} & 46.50 & 48.51{\scriptsize \textcolor[rgb]{0.2,0.6,0.2}{(+2.01)}} \\
        \bottomrule
    \end{tabular}
    }
    \vspace{-10pt}
\end{table}

\subsection{Main Results}
\begin{figure}[t]
    \centering
    \includegraphics[width=\linewidth]{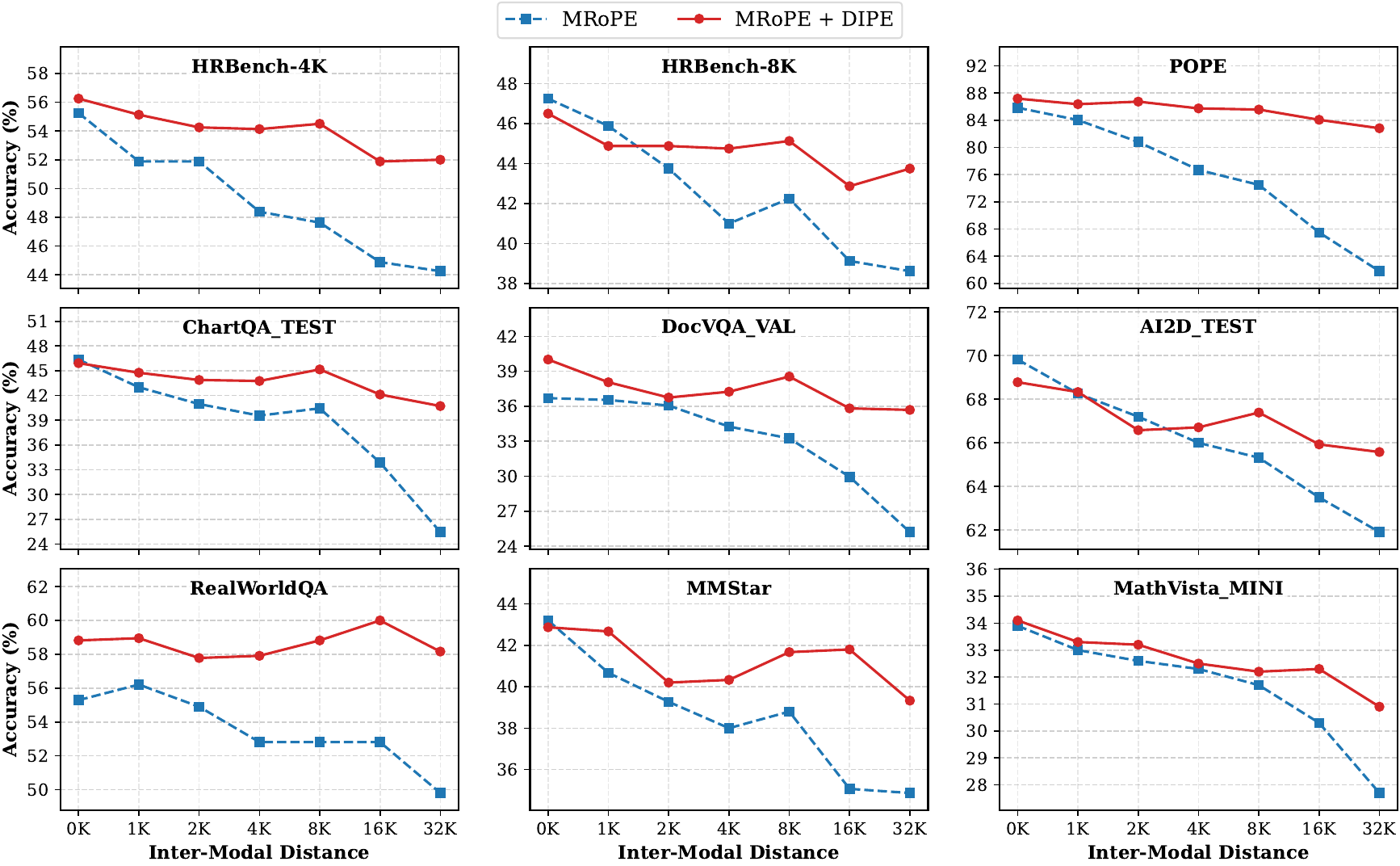}
    \vspace{-20pt}
    \caption{Accuracy across varying inter-modal distances. We compare the baseline MRoPE against MRoPE + DIPE on 9 benchmarks with distractor lengths ranging from 0K to 32K tokens.}
    \vspace{-15pt}
    \label{fig:acc_with_length}
\end{figure}

\noindent \textbf{Performance under the Long-Context VQA Protocol.}
To systematically evaluate robustness against visual fading, we assess the models from two complementary perspectives: a standardized extended-context evaluation (Table~\ref{tab:performance_evaluation}) and a continuous stress test across scaling context lengths (Fig.~\ref{fig:acc_with_length}).

To benchmark the overall capabilities across all 19 datasets, we establish a representative Long-Context VQA setting by introducing 8K tokens of textual distractors. Under this protocol, integrating DIPE consistently yields absolute average improvements of 2.00\%, 4.10\%, and 2.01\% for Vanilla RoPE, MRoPE and MRoPE-I, respectively.  While DIPE delivers consistent improvements across the vast majority of tasks, we also observe minor performance regressions on a few specific benchmarks. Nevertheless, considering the consistent overall improvements and the successful prevention of long-term visual fading, these isolated regressions are acceptable.

Notably, while the base MRoPE model exhibits lower overall accuracy than both Vanilla RoPE and MRoPE-I, it registers the most substantial gain (+4.10\%) when equipped with DIPE. As noted in recent literature~\cite{huang2025revisiting}, this initial baseline degradation stems from MRoPE's suboptimal frequency allocation across spatial dimensions, which inadvertently exacerbates the distance-induced decay in inter-modal attention. By explicitly anchoring the perceptual distance, DIPE effectively neutralizes this structural deficit.

To dynamically assess the progression of visual fading, we evaluate model performance across continuous distractor lengths ranging from 0K to 32K tokens. As illustrated in Fig.~\ref{fig:acc_with_length}, the baseline MRoPE suffers a severe and continuous decline in accuracy as the context scales. In contrast, MRoPE + DIPE maintains a remarkably stable performance trajectory. Crucially, the accuracy gap between the two configurations widens progressively as the sequence lengthens. This expanding margin compellingly demonstrates that DIPE effectively neutralizes the escalating inter-modal distance penalty, ensuring persistent visual grounding regardless of the generation length.

\begin{figure}[t]
    \centering
    \includegraphics[width=\textwidth]{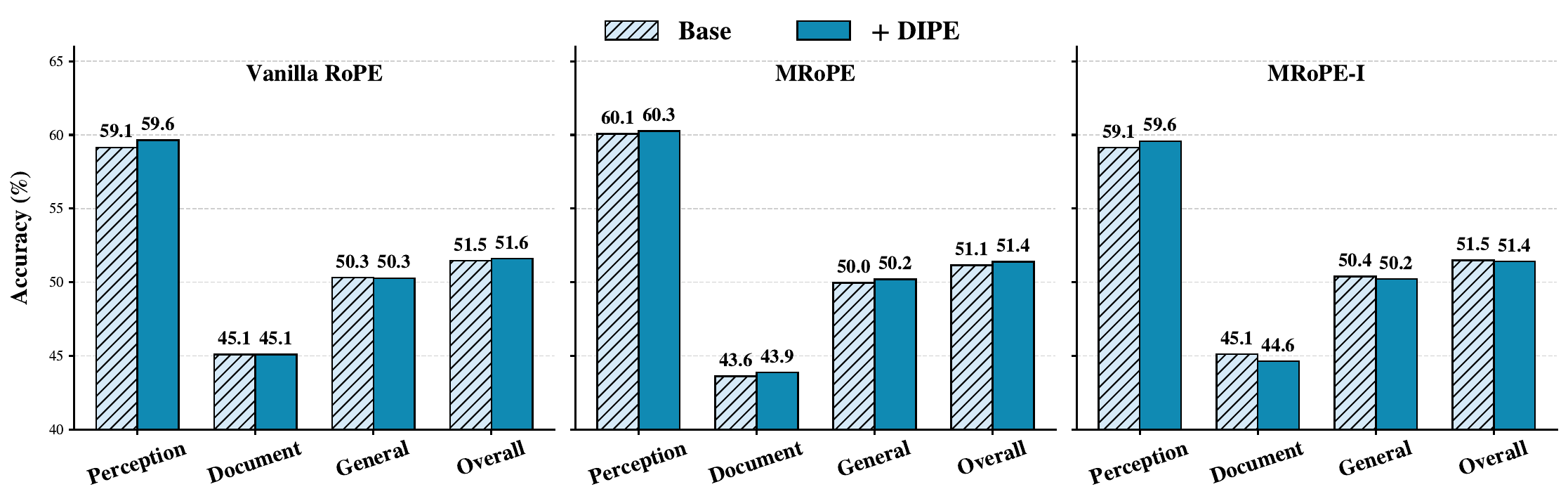}
    \vspace{-20pt}
    \caption{Performance on the Short-Context VQA protocol. DIPE serves as a non-destructive enhancement, maintaining the performance on standard VQA benchmarks.}
    \vspace{-15pt}
    \label{fig:shortvqa_results}
\end{figure}

\noindent \textbf{Performance under the Short-Context VQA Protocol.}
While the Long-Context VQA protocol explicitly highlights DIPE's capability to mitigate visual fading, it is equally critical to verify that this architectural modification operates as a non-destructive enhancement in standard short-context multimodal tasks. Fig.~\ref{fig:shortvqa_results} presents the results on the Short-Context VQA protocol (the complete results are provided in Appendix~\ref{appendix:shortvqa}). Quantitative evaluations reveal that DIPE-enhanced models maintain strict performance parity with their baseline counterparts across all three domains. For instance, MRoPE-I equipped with DIPE achieves an overall accuracy of 51.4\%, effectively matching the 51.5\% of the vanilla baseline. This stability remains consistent for both Vanilla RoPE and MRoPE architectures. These results empirically validate our core design principle: by orthogonally decoupling intra-modal and inter-modal geometries, DIPE preserves the essential 2D spatial locality of images and the 1D sequentiality of text in scenarios where visual fading is not yet a factor. This confirms DIPE as a non-destructive enhancement that mitigates visual fading while maintaining strict parity on standard benchmarks.

\begin{wrapfigure}{r}{0.44\textwidth}
    \centering
    \vspace{-20pt}
    \includegraphics[width=\linewidth]{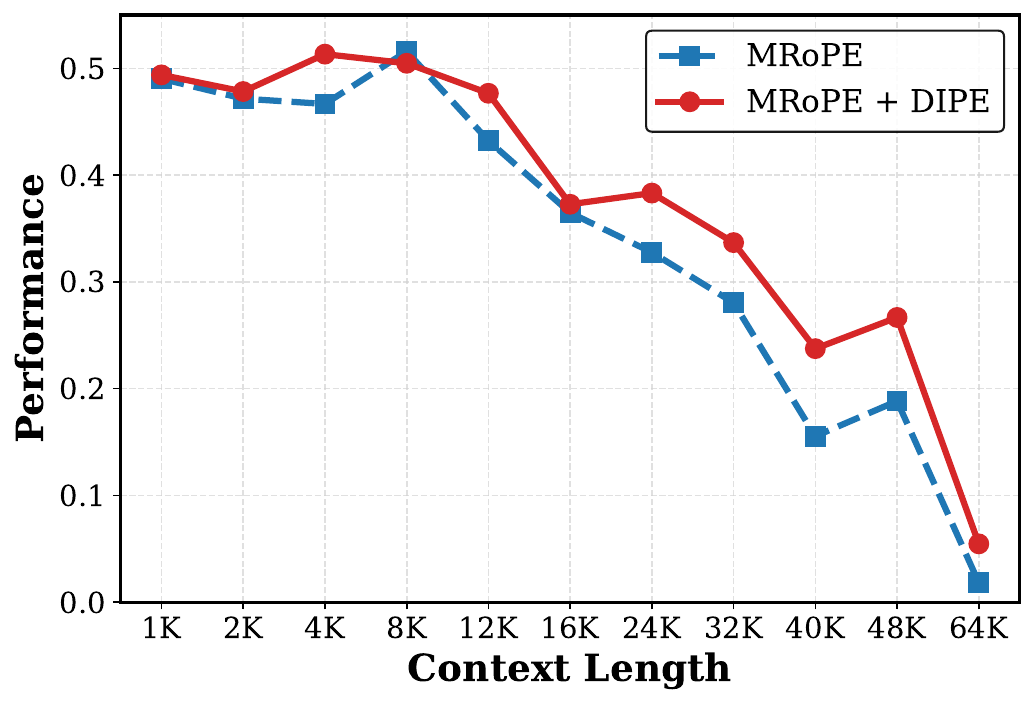}
    \vspace{-20pt}
    \caption{Performance on the image reasoning task in MM-NIAH~\cite{niah}.}
    \label{fig:mm_niah_reasoning}
    \vspace{-20pt}
\end{wrapfigure}

\noindent \textbf{Performance on the MM-NIAH Benchmark.}
To further validate the effectiveness of DIPE in standard long-context benchmark, we present the evaluation results on the image reasoning task from the MM-NIAH benchmark in Fig.~\ref{fig:mm_niah_reasoning}. While both configurations perform comparably at shorter context lengths, the standard MRoPE baseline exhibits a continuous performance degradation as the sequence scales. In contrast, the integration of DIPE effectively mitigates this decay, maintaining a significant performance margin over the baseline. Although both models inevitably experience a sharp accuracy drop at the extreme length of 64K tokens due to the inherent context capacity limits of the underlying language model, DIPE consistently demonstrates superior structural robustness in preserving visual reasoning capabilities.

\begin{table}[t]
    \centering
    \caption{Wall-clock latency under unpacked single-sequence evaluation. We report both full-model and attention-block latency.}
    \vspace{-10pt}
    \resizebox{\columnwidth}{!}{ 
    \begin{tabular}{llcccccc}
        \toprule
        \multicolumn{8}{c}{\textbf{Full Model (ms)}} \\
        \midrule
        Setting & Method & 1K & 2K & 4K & 8K & 16K & 32K \\
        \midrule
        \multirow{2}{*}{Forward}
        & MRoPE & 22.1$\pm$0.0 & 37.2$\pm$0.4 & 68.1$\pm$0.4 & 140.1$\pm$0.4 & 318.5$\pm$0.7 & 818.0$\pm$1.6 \\
        & +DIPE & 28.3$\pm$0.5 & 38.3$\pm$0.1 & 70.6$\pm$0.2 & 144.6$\pm$0.5 & 329.1$\pm$0.9 & 839.8$\pm$3.7 \\
        \midrule
        \multirow{2}{*}{Backward}
        & MRoPE & 43.0$\pm$0.2 & 69.8$\pm$0.4 & 135.4$\pm$0.5 & 293.0$\pm$0.8 & OOM & OOM \\
        & +DIPE & 47.9$\pm$0.5 & 79.4$\pm$0.1 & 150.6$\pm$0.2 & 322.0$\pm$1.0 & OOM & OOM \\
        \midrule
        \multirow{2}{*}{1-token Decoding}
        & MRoPE & 18.5$\pm$0.1 & 18.3$\pm$0.1 & 18.1$\pm$0.1 & 18.1$\pm$0.3 & 18.5$\pm$0.4 & 18.3$\pm$0.1 \\
        & +DIPE & 20.1$\pm$0.1 & 20.1$\pm$0.1 & 19.8$\pm$0.2 & 20.1$\pm$0.3 & 20.2$\pm$0.1 & 19.6$\pm$0.1 \\
        \midrule
        \multicolumn{8}{c}{\textbf{Attention Block (ms)}} \\
        \midrule
        Setting & Method & 1K & 2K & 4K & 8K & 16K & 32K \\
        \midrule
        \multirow{2}{*}{Forward}
        & MRoPE & 0.36$\pm$0.02 & 0.41$\pm$0.03 & 0.61$\pm$0.01 & 1.50$\pm$0.01 & 4.23$\pm$0.05 & 13.56$\pm$0.01 \\
        & +DIPE & 0.52$\pm$0.02 & 0.50$\pm$0.02 & 0.70$\pm$0.01 & 1.65$\pm$0.00 & 4.52$\pm$0.03 & 14.31$\pm$0.05 \\
        \midrule
        \multirow{2}{*}{Backward}
        & MRoPE & 0.83$\pm$0.01 & 1.23$\pm$0.02 & 1.49$\pm$0.01 & 3.62$\pm$0.02 & 10.88$\pm$0.03 & 37.29$\pm$0.85 \\
        & +DIPE & 1.52$\pm$0.02 & 1.54$\pm$0.03 & 2.19$\pm$0.01 & 4.51$\pm$0.03 & 12.46$\pm$0.06 & 40.22$\pm$0.34 \\
        \midrule
        \multirow{2}{*}{1-token Decoding}
        & MRoPE & 0.24$\pm$0.01 & 0.23$\pm$0.01 & 0.24$\pm$0.03 & 0.23$\pm$0.01 & 0.25$\pm$0.01 & 0.30$\pm$0.01 \\
        & +DIPE & 0.27$\pm$0.01 & 0.28$\pm$0.03 & 0.28$\pm$0.01 & 0.28$\pm$0.01 & 0.31$\pm$0.01 & 0.38$\pm$0.01 \\
        \bottomrule
    \end{tabular}
    }
    \label{tab:wallclock_efficiency}
\end{table}

\begin{wrapfigure}[9]{r}{0.49\textwidth}
    \centering
    \vspace{-25pt}
    \includegraphics[width=\linewidth]{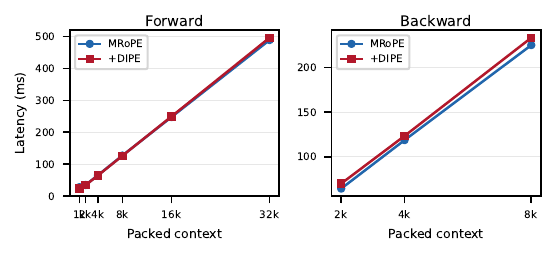}
    \vspace{-22pt}
    \caption{Training efficiency analysis with sequence packing.}
    \label{fig:packed_overhead}
    \vspace{-10pt}
\end{wrapfigure}

\noindent \textbf{Wall-Clock Efficiency Analysis.}
We further report wall-clock measurements in Table~\ref{tab:wallclock_efficiency}, covering forward, backward and single-token decoding latency. These measurements focus on the attention block, where the additional cost of DIPE is concentrated. DIPE introduces a modest overhead from split attention and LogSumExp fusion, while remaining close to the MRoPE baseline across long contexts. Under the common sequence-packing setting, the full-model latency exhibits a similar trend. As shown in Fig.~\ref{fig:packed_overhead}, MRoPE and +DIPE nearly overlap for packed training, indicating that the overhead remains small in standard training pipelines.

\begin{figure}[t]
    \centering
    \includegraphics[width=\linewidth]{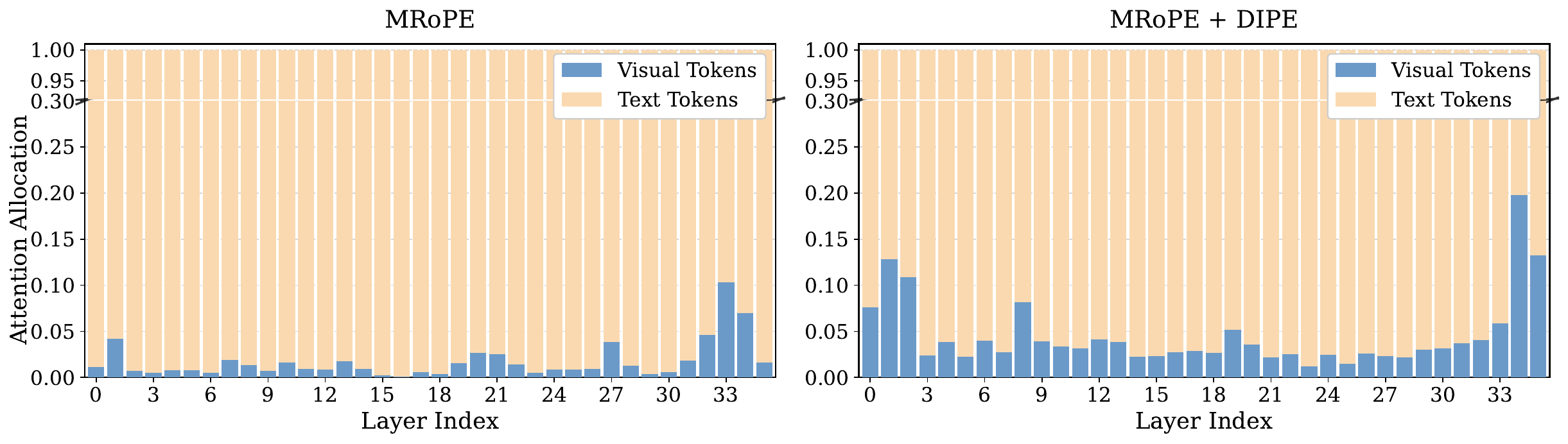}
    \vspace{-20pt}
    \caption{Layer-wise visual attention allocation analysis. The standard MRoPE baseline exhibits severe attention suppression, particularly in shallow layers. Integrating DIPE effectively restores this distribution across the network.}
    \vspace{-10pt}
    \label{fig:attention_allocate}
\end{figure}

\vspace{-3pt}
\subsection{In-depth Analysis}
\vspace{-2pt}
\noindent \textbf{Layer-Wise Visual Attention Analysis.}
To understand the underlying mechanism of how DIPE mitigates visual fading, we visualize the attention weights allocated to visual tokens. Fig.~\ref{fig:attention_allocate} illustrates the layer-wise attention allocation at a fixed distractor length of 8K tokens. Specifically, these weights are computed during the generation phase by averaging the attention scores from query tokens to visual tokens or text tokens across all heads. The standard MRoPE baseline exhibits severe suppression of visual attention, particularly within shallow layers. It indicates that early layers, heavily governed by spatial encodings, are highly vulnerable to distance-induced penalties. Although deeper layers retain marginal visual attention driven by semantic feature alignment, the visual signal remains significantly attenuated. By anchoring the inter-modal distance, DIPE restores the attention distribution across the network, rescuing positional perception in early layers and reinforcing visual constraints globally.

\begin{figure}[tbp]
    \centering
    \begin{minipage}{0.48\textwidth}
        \centering
        \includegraphics[width=\linewidth]{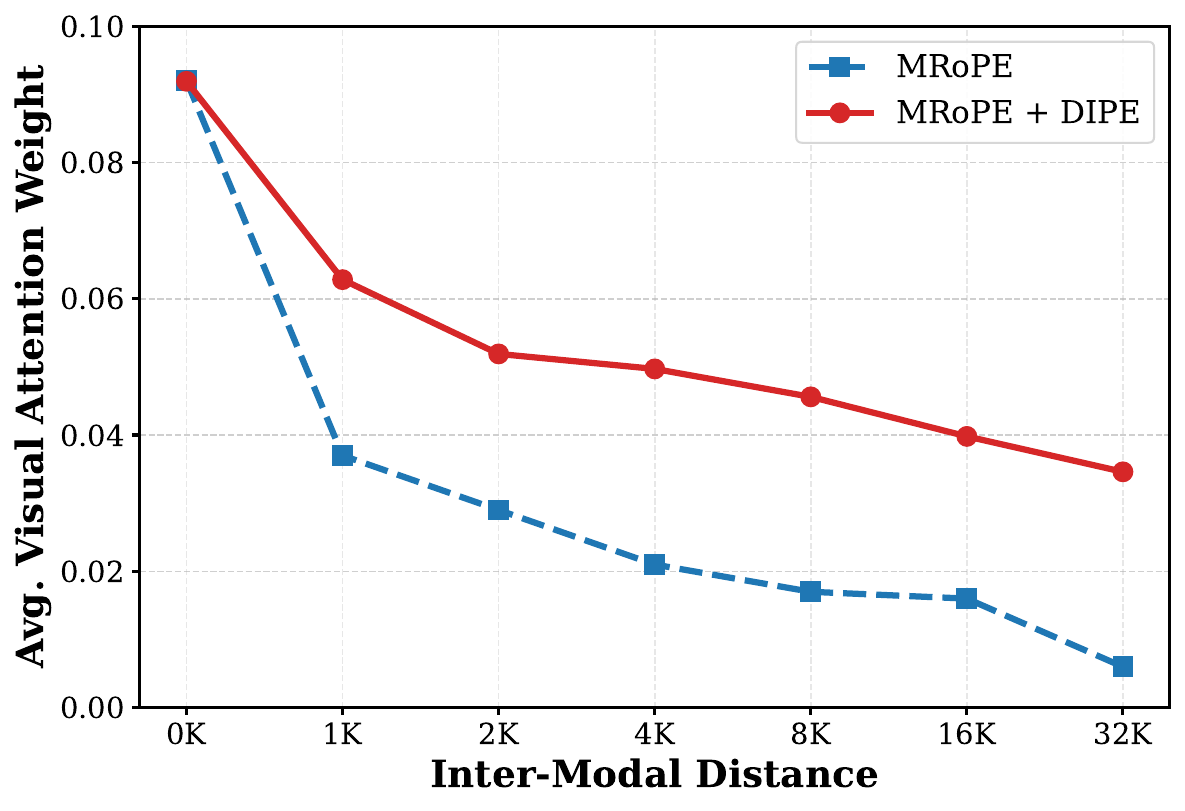}
        \vspace{-15pt}
        \caption{Average visual attention weight across inter-modal distances.}
        \label{fig:cmp_attention}
        \vspace{-5pt}
    \end{minipage}
    \hfill
    \begin{minipage}{0.48\textwidth}
        \centering
        \includegraphics[width=\linewidth]{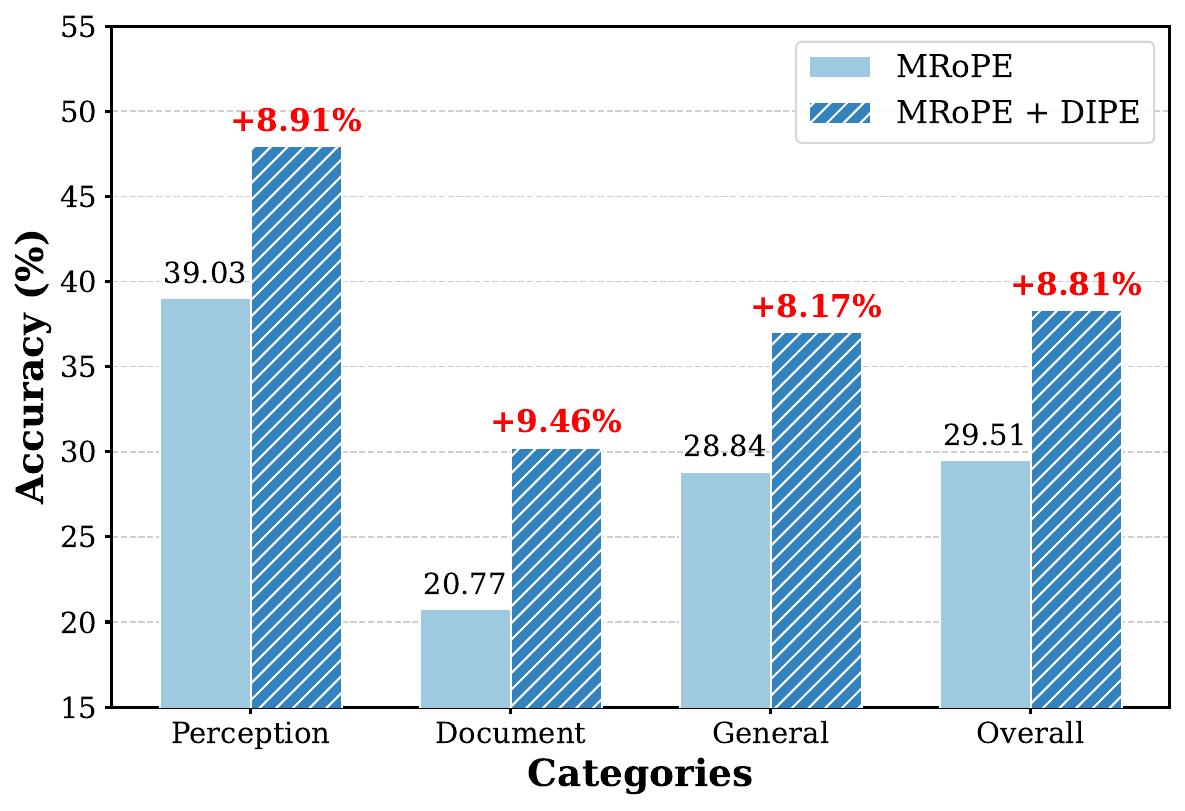}
        \vspace{-15pt}
        \caption{Performance comparison on Qwen2.5-0.5B as the language model.}
        \label{fig:model_scale}
        \vspace{-5pt}
    \end{minipage}
\end{figure}

\noindent \textbf{Mitigating Global Visual Attention Decay.}
To investigate the global trend of visual fading, we analyze the average visual attention weights across all layers under varying distractor lengths. As depicted in Fig.~\ref{fig:cmp_attention}, the baseline MRoPE demonstrates a rapid, monotonic decay in visual attention as the distractor length scales from 0K to 32K tokens, inherently detaching the model from visual constraints. In contrast, the integration of DIPE significantly mitigates this degradation trajectory. It is worth noting that a residual decline in attention persists even with DIPE. This is a natural phenomenon, since attention is inevitably diluted by the additional text tokens as the sequence length increases. However, by enforcing an invariant perceptual proximity, DIPE sustains substantially higher visual attention weights compared to the baseline, even at extreme inter-modal distances. Ultimately, this preserved visual grounding ensures that the model remains constrained by the image during extended generation, directly accounting for the superior accuracy observed in Long-Context VQA.

\begin{wraptable}{r}{0.44\textwidth}
    \vspace{-25pt}
    \centering
    \small
    \setlength{\tabcolsep}{6pt}
    \renewcommand{\arraystretch}{0.98}
    \caption{Performance comparison on Qwen3-1.7B as the language model.}
    \label{tab:generalization_analysis}
    \resizebox{\linewidth}{!}{ 
    \begin{tabular}{l cc}
        \toprule
        \textbf{Capability} & \textbf{Base} & \textbf{+DIPE} \\
        \midrule
        Perception & 50.11 & 54.92{\scriptsize \textcolor[rgb]{0.2,0.6,0.2}{(+4.81)}} \\
        Document & 35.34 & 36.56{\scriptsize \textcolor[rgb]{0.2,0.6,0.2}{(+1.22)}} \\
        General  & 35.90 & 42.25{\scriptsize \textcolor[rgb]{0.2,0.6,0.2}{(+6.35)}} \\
        \cmidrule{1-3}
        All & 40.21 & 44.46{\scriptsize \textcolor[rgb]{0.2,0.6,0.2}{(+4.25)}} \\
        \bottomrule
    \end{tabular}
    }
    \vspace{-17pt}
\end{wraptable}

\noindent \textbf{Generalization Across Model Architectures.} 
To verify the robustness of DIPE across different architectures, we conduct experiments on Qwen3-1.7B. As shown in Table~\ref{tab:generalization_analysis}, integrating DIPE yields consistent performance improvements across the evaluated domains. Specifically, the model achieves an absolute overall average gain of 4.25\%. These enhancements demonstrate that DIPE is a robust and effective solution for mitigating visual fading across different foundation models. The complete results are presented in Appendix~\ref{appendix:qwen3}.

\noindent \textbf{Generalization Across Model Scales.}
To investigate the generalization of DIPE across different scales, we evaluate it on the lightweight Qwen2.5-0.5B model. As shown in Fig.~\ref{fig:model_scale}, integrating DIPE delivers significant improvements across all domains, achieving an 8.81\% increase in overall accuracy. Notably, this gain is even larger than that observed in larger models. We attribute this to the fact that lightweight models inherently possess weaker long-context capabilities, rendering them more susceptible to visual fading. Detailed numerical results are provided in Appendix~\ref{appendix:scaling}.

\begin{wraptable}{r}{0.30\columnwidth}
    \vspace{-30pt}
    \centering
    \caption{DIPE maintains accuracy on interleaved tests, demonstrating its applicability to the interleaved data.}
    \label{tab:seq_benchmark}
    \resizebox{\linewidth}{!}{%
    \begin{tabular}{lc}
        \toprule
        \textbf{Model} & \textbf{Acc (\%)} \\
        \midrule
        Qwen2.5-3B-Instruct & 74.2 \\ \midrule
        MRoPE & 60.1 \\
        MRoPE + DIPE & 60.4 \\
        \bottomrule
    \end{tabular}%
    }
    \vspace{-15pt}
\end{wraptable}

\noindent \textbf{Robustness of DIPE in Image-Text Interleaved Contexts.}
\label{sec:interleave}
DIPE naturally accommodates interleaved image-text scenarios. Since the anchored position index is set to the start of each specific modality segment, distinct visual inputs inherently receive different temporal coordinates. This formulation guarantees that the global macroscopic order of the interleaved sequence is preserved. To verify that our decoupled encoding does not disrupt relative temporal positioning, we evaluated it on a custom 1K-sample interleaved test (Table~\ref{tab:seq_benchmark}). As expected, MRoPE+DIPE achieves performance strictly comparable to the baseline, confirming that DIPE successfully preserves sequential perception without degradation. Appendix~\ref{appendix:interleave} provides more details about this interleaved test. Furthermore, this robustness is also empirically validated on the MM-NIAH benchmark (Fig.~\ref{fig:mm_niah_reasoning}), which specifically targets long-context interleaved reasoning. Additionally, it also support multi-image scenarios and yields improvements on the multi-image BLINK benchmark (Table~\ref{tab:performance_evaluation}).

\vspace{-3pt}
\section{Conclusion}
This paper investigates the underlying mechanism of visual fading in MLLMs, revealing that the distance penalty inherent in MRoPE suppresses long-term inter-modal attention. To address this, we propose inter-modal Distance Invariant Position Encoding (DIPE), a simple but effective mechanism that decouples positional assignments based on modality interactions. It applies the sequential encoding to preserve intra-modal structures and the anchored encoding to anchor inter-modal distance. Extensive experiments demonstrate that DIPE effectively mitigates distance-induced visual fading in long-context scenarios, while preserving short-context accuracy. We hope our insights into inter-modal distance invariance inspire more robust position encoding designs for future MLLMs.

\section*{Acknowledgements}
This document is the results of the research project funded by the Strategic Priority Research Program of Chinese Academy of Sciences (Grant No. XDB0500103) and the National Natural Science Foundations of China (Grant No. 62306310). We would like to thank Jiazhen Liu for many fruitful discussions on the early formulation of our method and experimental design.

\bibliographystyle{splncs04}
\bibliography{main}

\title{Beyond Sequential Distance: Inter-Modal Distance Invariant Position Encoding \\ --------Supplementary Material--------}

\titlerunning{Inter-Modal Distance Invariant Position Encoding}
\author{ }
\authorrunning{L.~Chen et al.}
\institute{ }
\maketitle

This supplementary material contains additional details of the main manuscript and provides more experimental results. In Sec.~\ref{appendix:implementation_details}, we present the implementation details, including the training hyperparameters and evaluation protocols. Next, we provide the mathematical derivations of Eq.6 in Sec.~\ref{appendix:proof-lse}. Finally, we provide more experimental results in Sec.~\ref{appendix:additional_exp}.

\section{Implementation Details}
\label{appendix:implementation_details}

\subsection{Training Details}
\label{appendix:hyperparams}
As discussed in the main text, we adopt a two-stage training paradigm. The specific hyperparameter configurations for vision-language alignment (Stage 1) and multimodal instruction tuning (Stage 2) are summarized in Table~\ref{tab:hyperparams}.

\begin{table}[h]
    \centering
    \small
    \vspace{-10pt}
    \caption{Hyperparameter configurations for the training recipe.}
    \vspace{-5pt}
    \label{tab:hyperparams}
    \begin{tabular*}{\linewidth}{l@{\extracolsep{\fill}}cc}
    \toprule
    \textbf{Configuration} & \textbf{Stage 1} & \textbf{Stage 2} \\
    \midrule
    Trainable Modules & Projector & LLM + Projector \\
    Max Length & 8,192 & 8,192 \\
    Resolution & Native (Dynamic) & Native (Dynamic) \\
    Optimizer & AdamW & AdamW \\
    Batch Size & 256 & 128 \\
    Learning Rate & $1 \times 10^{-3}$ & $2 \times 10^{-5}$ \\
    LR Schedule & Cosine & Cosine \\
    Warmup Ratio & 0.03 & 0.03 \\
    Weight Decay & 0 & 0 \\
    Max Grad Norm & 1.0 & 1.0 \\
    Epochs & 1 & 1 \\
    Numerical Precision & BF16 & BF16 \\
    DeepSpeed Stage & ZeRO-2 & ZeRO-2 \\
    \bottomrule
    \end{tabular*}
    \vspace{-10pt}
\end{table}

\subsection{Evaluation Details}
\label{appendix:eval_details}
All benchmark evaluations are conducted using the open-source VLMEvalKit framework. To ensure strict reproducibility and eliminate decoding variance, we employ greedy decoding with the temperature set to $0.0$ and repetition penalty set to $1.0$. The evaluation prompts are rigorously aligned with the default configurations in VLMEvalKit.

\subsection{Details on the Interleaved Test}
\label{appendix:interleave}
\begin{figure}[h]
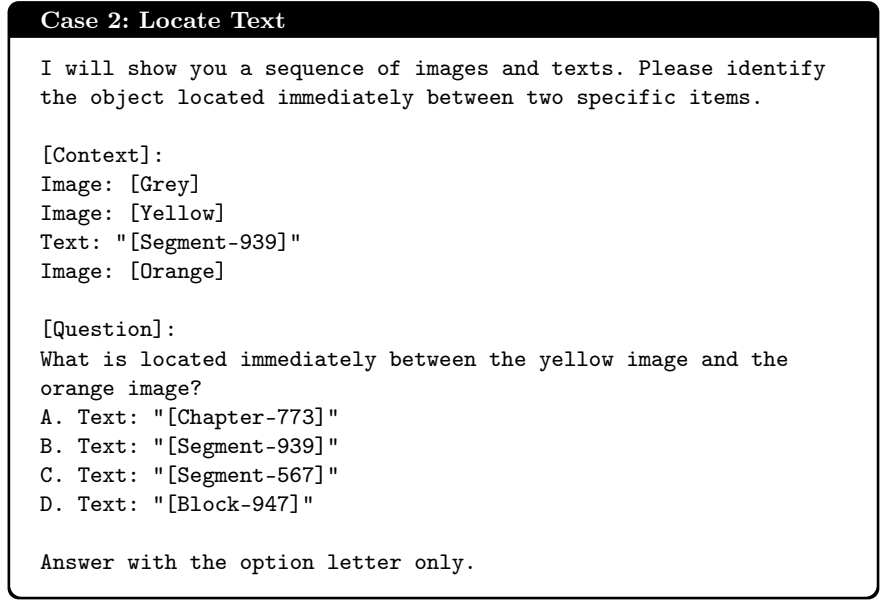

    \centering
    \begin{tcolorbox}[
        width=0.95\linewidth,
        title={Case 2: Locate Text},
        fonttitle=\bfseries,
        colframe=black,      
        colback=white,       
        colbacktitle=black,  
        coltitle=white,      
        fontupper=\small\ttfamily, 
        arc=1.5mm,             
        outer arc=1.5mm,       
        boxrule=1pt,         
        left=3mm,
        right=3mm
    ]
I will show you a sequence of images and texts. Please identify the object located immediately between two specific items.

\textbf{}

[Context]:

Image: [Grey]

Image: [Yellow]

Text: "[Segment-939]"

Image: [Orange]

\textbf{}

\textbf{[Question]:} 

What is located immediately between the yellow image and the orange image?

A. Text: "[Chapter-773]"

B. Text: "[Segment-939]"

C. Text: "[Segment-567]"

D. Text: "[Block-947]"

\textbf{}

Answer with the option letter only. 
    \end{tcolorbox}
    \caption{Example for identifying a text snippet within an interleaved sequence (1-shot example omitted for brevity).}
    \label{fig:case2}
\end{figure}

The core objective of this benchmark is to verify that our decoupled position encoding preserves the model's perception of relative image-text ordering. To evaluate this, we construct a 1K-sample synthetic interleaved benchmark. The context comprises a randomized sequence of uniformly colored image blocks (e.g., red, blue, green) and uniquely identifiable text markers (e.g., ``[Segment-156]''). Fundamentally, the benchmark probes the model's awareness of inter-modal relative positions by asking it to deduce the exact item located immediately between two specified context anchors. As illustrated in Fig.~\ref{fig:case1} and Fig.~\ref{fig:case2}, we specifically design two types of evaluation scenarios to comprehensively assess this capability: (1) locating a target image bounded by surrounding anchors, and (2) locating target text bounded by surrounding anchors. Crucially, our proposed DIPE configuration maintains strict performance parity with the baseline on this benchmark. This confirms that explicitly anchoring the inter-modal perceptual distance does not compromise the model's fundamental ability to track the relative sequence of interleaved modalities.

\begin{figure}[h]
    \centering
    \begin{tcolorbox}[
        width=0.95\linewidth,
        title={Case 1: Locate Image},
        fonttitle=\bfseries,
        colframe=black,      
        colback=white,       
        colbacktitle=black,  
        coltitle=white,      
        fontupper=\small\ttfamily, 
        arc=1.5mm,             
        outer arc=1.5mm,       
        boxrule=1pt,         
        left=3mm,
        right=3mm
    ]
I will show you a sequence of images and texts. Please identify the object located immediately between two specific items.

\textbf{}

[Context]:

Image: [Orange]

Image: [Green]

Text: "[Segment-582]"

Image: [Red]

\textbf{}

\textbf{[Question]:} 

What is located immediately between the orange image and the text snippet "[Segment-582]"?

A. Green image

B. Black image

C. Grey image

D. Red image

\textbf{}

Answer with the option letter only. 
    \end{tcolorbox}
    \caption{Example for identifying an image within an interleaved sequence (1-shot example omitted for brevity).}
    \label{fig:case1}
\end{figure}

\vspace{-10pt}
\section{Derivation of Eq.6}
\label{appendix:proof-lse}
In the main text, we present a method to merge the outputs of the decoupled intra-modal and inter-modal attention kernels using the LogSumExp (LSE) statistic. We derive this numerically stable fusion here.

Standard attention requires a global softmax normalization over all keys. Let the unnormalized attention score between query $i$ and key $j$ be $s_{ij}$. The mathematically exact global output $\mathbf{O}$ requires the global partition function $Z$:
\begin{equation}
    Z = \sum_{j \in \text{intra}} \exp(s_{1j}) + \sum_{k \in \text{inter}} \exp(s_{2k}).
\end{equation}

When the attention kernels compute the split partitions, they return the partial outputs $\mathbf{O}_1, \mathbf{O}_2$ and their corresponding LSE values $\ell_1, \ell_2$, where $\ell_1 = \log \sum_{j \in \text{intra}} \exp(s_{1j})$ and $\ell_2 = \log \sum_{k \in \text{inter}} \exp(s_{2k})$. Therefore, the global denominator can be perfectly reconstructed as $Z = e^{\ell_1} + e^{\ell_2}$.

The globally normalized output is the sum of the unnormalized partial outputs divided by $Z$:
\begin{equation}
    \mathbf{O} = \frac{e^{\ell_1} \mathbf{O}_1 + e^{\ell_2} \mathbf{O}_2}{e^{\ell_1} + e^{\ell_2}}.
\end{equation}

To avoid numerical instability and eliminate the need for direct exponentiation, we can rewrite the weighting coefficients using the standard sigmoid function $\sigma(x) = \frac{1}{1 + e^{-x}}$. By dividing the numerator and denominator by $e^{\ell_1}$ and $e^{\ell_2}$ respectively, we get:
\begin{align}
    \frac{e^{\ell_1}}{e^{\ell_1} + e^{\ell_2}} &= \frac{1}{1 + e^{\ell_2 - \ell_1}} = \sigma(\ell_1 - \ell_2), \\
    \frac{e^{\ell_2}}{e^{\ell_1} + e^{\ell_2}} &= \frac{1}{1 + e^{\ell_1 - \ell_2}} = \sigma(\ell_2 - \ell_1).
\end{align}

Substituting these into the sum yields the final fusion equation used in our implementation (Eq.~6):
\begin{equation}
    \mathbf{O} = \sigma(\ell_1 - \ell_2) \mathbf{O}_1 + \sigma(\ell_2 - \ell_1) \mathbf{O}_2,
\end{equation}
where $\sigma(\ell_1 - \ell_2) + \sigma(\ell_2 - \ell_1) = 1$.

\section{Additional Experiments Results}
\label{appendix:additional_exp}

\subsection{Complete Results on Short-Context VQA}
\label{appendix:shortvqa}
Fig.~\ref{fig:shortvqa_results} in the main manuscript summarizes our evaluations under the Short-Context VQA protocol. Table~\ref{tab:performance_evaluation_short} provides the complete, detailed quantitative results across all 19 benchmarks. These results comprehensively demonstrate that integrating DIPE preserves fundamental multimodal capabilities and maintains strict performance parity with standard baselines in short-context scenarios.

\subsection{Complete Results on Generalization Analysis}
\label{appendix:qwen3}
Table~\ref{tab:qwen3_complete} presents the comprehensive evaluation results of our proposed DIPE mechanism when applied to the Qwen3-1.7B architecture. The results demonstrate that DIPE consistently improves performance on the vast majority of tasks. We observe particularly notable gains in tasks requiring extensive context processing and fine-grained visual reasoning, such as RealWorldQA (+17.12\%), POPE (+11.67\%), and BLINK (+9.94\%). While minor performance fluctuations are observed in a few document understanding tasks (e.g., ChartQA and OCRBench), the overall average performance across all three domains exhibits a solid and consistent upward trend.

\subsection{Complete Results on Scaling Analysis}
\label{appendix:scaling}
Table~\ref{tab:qwen2_5_evaluation} presents the detailed quantitative results for the scaling analysis, comparing the lightweight Qwen2.5-0.5B model with the Qwen2.5-3B baseline. The results demonstrate that integrating DIPE yields consistent and substantial improvements across all three domains. Notably, the 0.5B model achieves an overall accuracy gain of 8.81\%, which is even more pronounced than the improvement observed in the 3B model.

\begin{table}[t]
    \centering
    \small
    \setlength{\tabcolsep}{3pt}
    \renewcommand{\arraystretch}{1.05}
    \caption{Complete results under the Short-Context VQA protocol. We systematically compare the Vanilla RoPE, MRoPE and MRoPE-I baselines both with and without our proposed DIPE mechanism. It shows that DIPE does not compromise the capability on standard short-context scenarios.}
    \vspace{-8pt}
    \label{tab:performance_evaluation_short}
    \resizebox{\columnwidth}{!}{ 
    \begin{tabular}{ll cccccc}
        \toprule
        \multirow{2}{*}{Capability} & \multirow{2}{*}{Benchmark} & \multicolumn{2}{c}{Vanilla RoPE} & \multicolumn{2}{c}{MRoPE} & \multicolumn{2}{c}{MRoPE-I} \\
        \cmidrule(lr){3-4} \cmidrule(lr){5-6} \cmidrule(lr){7-8}
        & & Base & +DIPE & Base & +DIPE & Base & +DIPE \\
        \midrule
        
        \multirow{6}{*}{Perception} 
        & HRBench-4K & 53.12 & 53.62 & 55.25 & 56.25 & 54.75 & 56.25 \\
        & HRBench-8K & 47.38 & 48.00 & 47.25 & 46.50 & 44.25 & 44.12 \\
        & V$^*$ & 49.21 & 49.74 & 51.31 & 48.69 & 51.31 & 50.26 \\
        & POPE & 85.52 & 85.70 & 85.85 & 87.19 & 85.80 & 86.08 \\
        & CountBench & 76.99 & 77.60 & 77.80 & 80.65 & 75.56 & 78.62 \\
        & BLINK$_{\mathrm{val}}$ & 42.66 & 43.19 & 43.03 & 42.35 & 43.14 & 42.14 \\
        \midrule

        \multirow{6}{*}{\shortstack{Document \\ Understanding}} 
        & ChartQA$_{\mathrm{test}}$ & 48.24 & 48.40 & 46.36 & 45.92 & 48.88 & 47.56 \\
        & DocVQA$_{\mathrm{val}}$ & 43.54 & 43.22 & 36.70 & 40.01 & 41.72 & 41.22 \\
        & TextVQA$_{\mathrm{val}}$ & 57.36 & 56.28 & 57.52 & 57.27 & 56.68 & 56.12 \\
        & AI2D$_{\mathrm{test}}$ & 69.11 & 69.62 & 69.82 & 68.78 & 70.63 & 69.40 \\
        & InfoVQA$_{\mathrm{val}}$ & 23.80 & 24.10 & 23.15 & 22.47 & 24.28 & 24.79 \\
        & OCRBench & 28.50 & 29.00 & 28.10 & 28.80 & 28.50 & 28.70 \\
        \midrule

        \multirow{7}{*}{General VQA}
        & RealWorldQA & 56.21 & 57.52 & 55.29 & 58.82 & 56.47 & 56.86 \\
        & MMStar & 42.27 & 41.73 & 43.20 & 42.87 & 42.73 & 41.93 \\
        & MMBenchV1.1-EN$_{\mathrm{dev}}$ & 67.49 & 67.96 & 69.04 & 67.88 & 66.33 & 67.26 \\
        & MMBenchV1.1-CN$_{\mathrm{dev}}$ & 66.25 & 66.25 & 66.95 & 66.02 & 65.94 & 65.48 \\
        & MMVP & 63.67 & 64.00 & 61.00 & 63.00 & 66.33 & 64.33 \\
        & MathVision$_{\mathrm{mini}}$ & 22.37 & 20.39 & 21.71 & 19.41 & 21.05 & 21.38 \\
        & MathVista$_{\mathrm{mini}}$ & 33.90 & 34.10 & 32.60 & 33.30 & 33.90 & 34.30 \\ \midrule

        \multirow{5}{*}{Overall} 
        & Perception & 59.15 & 59.64 & 60.08 & 60.27 & 59.14 & 59.58 \\
        & Document & 45.09 & 45.10 & 43.61 & 43.88 & 45.12 & 44.63 \\
        & General & 50.31 & 50.28 & 49.97 & 50.19 & 50.39 & 50.22 \\
        \cmidrule{2-8}
        & All & 51.45 & 51.60 & 51.15 & 51.38 & 51.49 & 51.41 \\
        \bottomrule
    \end{tabular}
    }
    \vspace{-5pt}
\end{table}

\begin{table}[t]
    \centering
    \small
    \setlength{\tabcolsep}{8pt}
    \renewcommand{\arraystretch}{1.05}
    \caption{Complete results of the Qwen3-1.7B model under the Long-Context VQA protocol. Green numbers indicate performance improvements, while red numbers indicate performance drops after integrating DIPE.}
    \vspace{-8pt}
    \label{tab:qwen3_complete}
    \begin{tabular}{ll ll}
        \toprule
        \multirow{2}{*}{Capability} & \multirow{2}{*}{Benchmark} & \multicolumn{2}{c}{MRoPE} \\
        \cmidrule(lr){3-4}
        & & Base & +DIPE \\
        \midrule
        
        \multirow{6}{*}{Perception} 
        & HRBench-4K & 42.88 & 44.62{\scriptsize \textcolor[rgb]{0.2,0.6,0.2}{(+1.74)}} \\
        & HRBench-8K & 36.62 & 37.00{\scriptsize \textcolor[rgb]{0.2,0.6,0.2}{(+0.38)}} \\
        & V$^*$ & 48.17 & 51.31{\scriptsize \textcolor[rgb]{0.2,0.6,0.2}{(+3.14)}} \\
        & POPE & 66.95 & 78.62{\scriptsize \textcolor[rgb]{0.2,0.6,0.2}{(+11.67)}} \\
        & CountBench & 74.34 & 76.37{\scriptsize \textcolor[rgb]{0.2,0.6,0.2}{(+2.03)}} \\
        & BLINK$_{\mathrm{val}}$ & 31.67 & 41.61{\scriptsize \textcolor[rgb]{0.2,0.6,0.2}{(+9.94)}} \\
        \midrule

        \multirow{6}{*}{\shortstack{Document \\ Understanding}} 
        & ChartQA$_{\mathrm{test}}$ & 41.92 & 40.68{\scriptsize \textcolor[rgb]{0.8,0.2,0.2}{(-1.24)}} \\
        & DocVQA$_{\mathrm{val}}$ & 28.33 & 32.44{\scriptsize \textcolor[rgb]{0.2,0.6,0.2}{(+4.11)}} \\
        & TextVQA$_{\mathrm{val}}$ & 41.67 & 45.43{\scriptsize \textcolor[rgb]{0.2,0.6,0.2}{(+3.76)}} \\
        & AI2D$_{\mathrm{test}}$ & 62.27 & 62.11{\scriptsize \textcolor[rgb]{0.8,0.2,0.2}{(-0.16)}} \\
        & InfoVQA$_{\mathrm{val}}$ & 16.15 & 18.19{\scriptsize \textcolor[rgb]{0.2,0.6,0.2}{(+2.04)}} \\
        & OCRBench & 21.70 & 20.50{\scriptsize \textcolor[rgb]{0.8,0.2,0.2}{(-1.20)}} \\
        \midrule

        \multirow{7}{*}{General VQA}
        & RealWorldQA & 31.90 & 49.02{\scriptsize \textcolor[rgb]{0.2,0.6,0.2}{(+17.12)}} \\
        & MMStar & 35.47 & 36.47{\scriptsize \textcolor[rgb]{0.2,0.6,0.2}{(+1.00)}} \\
        & MMBenchV1.1-EN$_{\mathrm{dev}}$ & 46.83 & 55.19{\scriptsize \textcolor[rgb]{0.2,0.6,0.2}{(+8.36)}} \\
        & MMBenchV1.1-CN$_{\mathrm{dev}}$ & 44.43 & 53.17{\scriptsize \textcolor[rgb]{0.2,0.6,0.2}{(+8.74)}} \\
        & MMVP & 52.00 & 58.00{\scriptsize \textcolor[rgb]{0.2,0.6,0.2}{(+6.00)}} \\
        & MathVision$_{\mathrm{mini}}$ & 13.16 & 16.12{\scriptsize \textcolor[rgb]{0.2,0.6,0.2}{(+2.96)}} \\
        & MathVista$_{\mathrm{mini}}$ & 27.50 & 27.80{\scriptsize \textcolor[rgb]{0.2,0.6,0.2}{(+0.30)}} \\ \midrule

        \multirow{4}{*}{Overall} 
        & Perception & 50.11 & 54.92{\scriptsize \textcolor[rgb]{0.2,0.6,0.2}{(+4.81)}} \\
        & Document & 35.34 & 36.56{\scriptsize \textcolor[rgb]{0.2,0.6,0.2}{(+1.22)}} \\
        & General & 35.90 & 42.25{\scriptsize \textcolor[rgb]{0.2,0.6,0.2}{(+6.35)}} \\
        \cmidrule{2-4}
        & All & 40.21 & 44.46{\scriptsize \textcolor[rgb]{0.2,0.6,0.2}{(+4.25)}} \\
        \bottomrule
    \end{tabular}
    \vspace{-5pt}
\end{table}

\begin{table}[t!]
    \centering
    \small
    \setlength{\tabcolsep}{3pt}
    \renewcommand{\arraystretch}{1.03}
    \caption{Performance evaluation of Qwen2.5-0.5B and Qwen2.5-3B models. Green numbers indicate performance improvements, while red numbers indicate performance drops after integrating DIPE.}
    \vspace{-5pt}
    \label{tab:qwen2_5_evaluation}
    \resizebox{\columnwidth}{!}{ 
    \begin{tabular}{ll llll}
        \toprule
        \multirow{2}{*}{\textbf{Capability}} & \multirow{2}{*}{\textbf{Benchmark}} & \multicolumn{2}{c}{\textbf{Qwen2.5-0.5B}} & \multicolumn{2}{c}{\textbf{Qwen2.5-3B}} \\
        \cmidrule(lr){3-4} \cmidrule(lr){5-6}
        & & Base & +DIPE & Base & +DIPE \\
        \midrule
        
        \multirow{6}{*}{Perception} 
        & CountBench & 30.35 & 63.95{\scriptsize \textcolor[rgb]{0.2,0.6,0.2}{(+33.60)}} & 66.60 & 75.97{\scriptsize \textcolor[rgb]{0.2,0.6,0.2}{(+9.37)}} \\
        & HRBench-4K & 34.13 & 37.25{\scriptsize \textcolor[rgb]{0.2,0.6,0.2}{(+3.12)}} & 47.63 & 54.50{\scriptsize \textcolor[rgb]{0.2,0.6,0.2}{(+6.87)}} \\
        & HRBench-8K & 30.00 & 31.63{\scriptsize \textcolor[rgb]{0.2,0.6,0.2}{(+1.63)}} & 42.25 & 45.13{\scriptsize \textcolor[rgb]{0.2,0.6,0.2}{(+2.88)}} \\
        & V$^*$ & 36.13 & 36.65{\scriptsize \textcolor[rgb]{0.2,0.6,0.2}{(+0.52)}} & 49.21 & 50.26{\scriptsize \textcolor[rgb]{0.2,0.6,0.2}{(+1.05)}} \\
        & POPE & 67.87 & 81.77{\scriptsize \textcolor[rgb]{0.2,0.6,0.2}{(+13.90)}} & 74.50 & 85.58{\scriptsize \textcolor[rgb]{0.2,0.6,0.2}{(+11.08)}} \\
        & BLINK$_{\mathrm{val}}$ & 35.72 & 36.40{\scriptsize \textcolor[rgb]{0.2,0.6,0.2}{(+0.68)}} & 39.87 & 41.93{\scriptsize \textcolor[rgb]{0.2,0.6,0.2}{(+2.06)}} \\
        \midrule

        \multirow{6}{*}{\shortstack{Document \\ Understanding}} 
        & AI2D$_{\mathrm{test}}$ & 48.87 & 51.62{\scriptsize \textcolor[rgb]{0.2,0.6,0.2}{(+2.75)}} & 65.32 & 67.39{\scriptsize \textcolor[rgb]{0.2,0.6,0.2}{(+2.07)}} \\
        & ChartQA$_{\mathrm{test}}$ & 13.20 & 32.64{\scriptsize \textcolor[rgb]{0.2,0.6,0.2}{(+19.44)}} & 40.44 & 45.16{\scriptsize \textcolor[rgb]{0.2,0.6,0.2}{(+4.72)}} \\
        & TextVQA$_{\mathrm{val}}$ & 25.06 & 41.83{\scriptsize \textcolor[rgb]{0.2,0.6,0.2}{(+16.77)}} & 46.61 & 50.03{\scriptsize \textcolor[rgb]{0.2,0.6,0.2}{(+3.42)}} \\
        & InfoVQA$_{\mathrm{val}}$ & 13.67 & 13.51{\scriptsize \textcolor{red}{(-0.16)}} & 23.60 & 21.84{\scriptsize \textcolor{red}{(-1.76)}} \\
        & DocVQA$_{\mathrm{val}}$ & 15.84 & 27.16{\scriptsize \textcolor[rgb]{0.2,0.6,0.2}{(+11.32)}} & 33.25 & 38.55{\scriptsize \textcolor[rgb]{0.2,0.6,0.2}{(+5.30)}} \\
        & OCRBench & 8.00 & 14.60{\scriptsize \textcolor[rgb]{0.2,0.6,0.2}{(+6.60)}} & 22.50 & 23.40{\scriptsize \textcolor[rgb]{0.2,0.6,0.2}{(+0.90)}} \\
        \midrule

        \multirow{7}{*}{General VQA}
        & MathVision$_{\mathrm{mini}}$ & 15.46 & 15.79{\scriptsize \textcolor[rgb]{0.2,0.6,0.2}{(+0.33)}} & 18.09 & 19.08{\scriptsize \textcolor[rgb]{0.2,0.6,0.2}{(+0.99)}} \\
        & MathVista$_{\mathrm{mini}}$ & 23.80 & 26.10{\scriptsize \textcolor[rgb]{0.2,0.6,0.2}{(+2.30)}} & 31.70 & 32.20{\scriptsize \textcolor[rgb]{0.2,0.6,0.2}{(+0.50)}} \\
        & RealWorldQA & 41.05 & 43.92{\scriptsize \textcolor[rgb]{0.2,0.6,0.2}{(+2.87)}} & 52.81 & 58.82{\scriptsize \textcolor[rgb]{0.2,0.6,0.2}{(+6.01)}} \\
        & MMStar & 28.33 & 34.00{\scriptsize \textcolor[rgb]{0.2,0.6,0.2}{(+5.67)}} & 38.80 & 41.67{\scriptsize \textcolor[rgb]{0.2,0.6,0.2}{(+2.87)}} \\
        & MMBenchV1.1-EN$_{\mathrm{dev}}$ & 23.68 & 46.52{\scriptsize \textcolor[rgb]{0.2,0.6,0.2}{(+22.84)}} & 50.00 & 58.75{\scriptsize \textcolor[rgb]{0.2,0.6,0.2}{(+8.75)}} \\
        & MMBenchV1.1-CN$_{\mathrm{dev}}$ & 19.20 & 39.71{\scriptsize \textcolor[rgb]{0.2,0.6,0.2}{(+20.51)}} & 49.30 & 56.66{\scriptsize \textcolor[rgb]{0.2,0.6,0.2}{(+7.36)}} \\
        & MMVP & 50.33 & 53.00{\scriptsize \textcolor[rgb]{0.2,0.6,0.2}{(+2.67)}} & 56.67 & 60.00{\scriptsize \textcolor[rgb]{0.2,0.6,0.2}{(+3.33)}} \\
        \midrule

        \multirow{5}{*}{Overall} 
        & Perception & 39.03 & 47.94{\scriptsize \textcolor[rgb]{0.2,0.6,0.2}{(+8.91)}} & 53.34 & 58.90{\scriptsize \textcolor[rgb]{0.2,0.6,0.2}{(+5.56)}} \\
        & Document & 20.77 & 30.23{\scriptsize \textcolor[rgb]{0.2,0.6,0.2}{(+9.46)}} & 38.62 & 41.06{\scriptsize \textcolor[rgb]{0.2,0.6,0.2}{(+2.44)}} \\
        & General & 28.84 & 37.01{\scriptsize \textcolor[rgb]{0.2,0.6,0.2}{(+8.17)}} & 42.48 & 46.74{\scriptsize \textcolor[rgb]{0.2,0.6,0.2}{(+4.26)}} \\
        \cmidrule{2-6}
        & All & 29.51 & 38.32{\scriptsize \textcolor[rgb]{0.2,0.6,0.2}{(+8.81)}} & 44.69 & 48.79{\scriptsize \textcolor[rgb]{0.2,0.6,0.2}{(+4.10)}} \\
        \bottomrule
    \end{tabular}
    }
    \vspace{-5pt}
\end{table}

\end{document}